\documentclass[lettersize,journal]{IEEEtran}
\usepackage{amsmath,amsfonts}
\usepackage{algorithmic}
\usepackage{array}

\usepackage{textcomp}
\usepackage{stfloats}
\usepackage{url}
\usepackage{verbatim}
\usepackage{graphicx}
\usepackage{bm}
\usepackage{algorithm}
\usepackage{multirow}
\usepackage{color}
\usepackage{subfigure}

\usepackage[english]{babel}
\usepackage{amsthm}

\newtheorem{proposition}{Proposition}

\def\BibTeX{{\rm B\kern-.05em{\sc i\kern-.025em b}\kern-.08em
    T\kern-.1667em\lower.7ex\hbox{E}\kern-.125emX}}

\begin{document}
\title{Robust Graph Matching Using An Unbalanced Hierarchical Optimal Transport Framework}
\author{Haoran Cheng, Dixin Luo, Hongteng Xu~\IEEEmembership{Member,~IEEE}
\thanks{Manuscript created October, 2024. This work was supported in part by the National Natural Science Foundation of China (62102031, 62106271, 92270110), and the foundation of Key Laboratory of Artificial Intelligence, Ministry of Education, China. (Corresponding author: Dixin Luo.)

Haoran Cheng is with the School of Computer Science and Technology, Beijing Institute of Technology, Beijing 100081, China (e-mail: haoran.cheng@bit.edu.cn).

Dixin Luo is with the School of Computer Science and Technology, Beijing Institute of Technology, Beijing 100081, China, and also with the Key Laboratory of Artificial Intelligence, Ministry of Education, China (e-mail: dixin.luo@bit.edu.cn).

Hongteng Xu is with the Gaoling School of Artificial Intelligence, Renmin University of China, Beijing, China, and also with the Beijing Key Laboratory of Big Data Management and Analysis Methods, China (e-mail: hongtengxu@ruc.edu.cn).
}}

\markboth{Journal of \LaTeX\ Class Files,~Vol.~18, No.~9, September~2020}%
{How to Use the IEEEtran \LaTeX \ Templates}

\maketitle

\begin{abstract}
Graph matching is one of the most significant graph analytic tasks, which aims to find the node correspondence across different graphs. 
Most existing graph matching approaches mainly rely on topological information, whose performances are often sub-optimal and sensitive to data noise because of not fully leveraging the multi-modal information hidden in graphs, such as node attributes, subgraph structures, etc. 
In this study, we propose a novel and robust graph matching method based on an unbalanced hierarchical optimal transport (UHOT) framework, which, to our knowledge, makes the first attempt to exploit cross-modal alignment in graph matching. 
In principle, applying multi-layer message passing, we represent each graph as layer-wise node embeddings corresponding to different modalities. 
Given two graphs, we align their node embeddings within the same modality and across different modalities, respectively. 
Then, we infer the node correspondence by the weighted average of all the alignment results.
This method is implemented as computing the UHOT distance between the two graphs --- each alignment is achieved by a node-level optimal transport plan between two sets of node embeddings, and the weights of all alignment results correspond to an unbalanced modality-level optimal transport plan. 
Experiments on various graph matching tasks demonstrate the superiority and robustness of our method compared to state-of-the-art approaches. 
Our implementation is available at \url{https://github.com/Dixin-Lab/UHOT-GM}.
\end{abstract}


\section{Introduction}\label{sec:intro}
Graph matching aims to find the node correspondence across different graphs, which commonly appears in many practical applications. 
For instance, protein-protein interaction (PPI) network alignment~\cite{singh2008global,liu2017novel} helps to explore the functionally-similar proteins of different species. 
Linking user accounts in different social networks benefits personalized recommendation~\cite{li2019partially,li2018distribution} and fraud detection~\cite{huang2022auc,hooi2017graph}. 
Vision tasks like shape matching can be formulated as graph matching problems~\cite{vento2013graph,fey2020deep}.

\begin{figure}[t]
    \centering
    \includegraphics[width=0.95\linewidth]{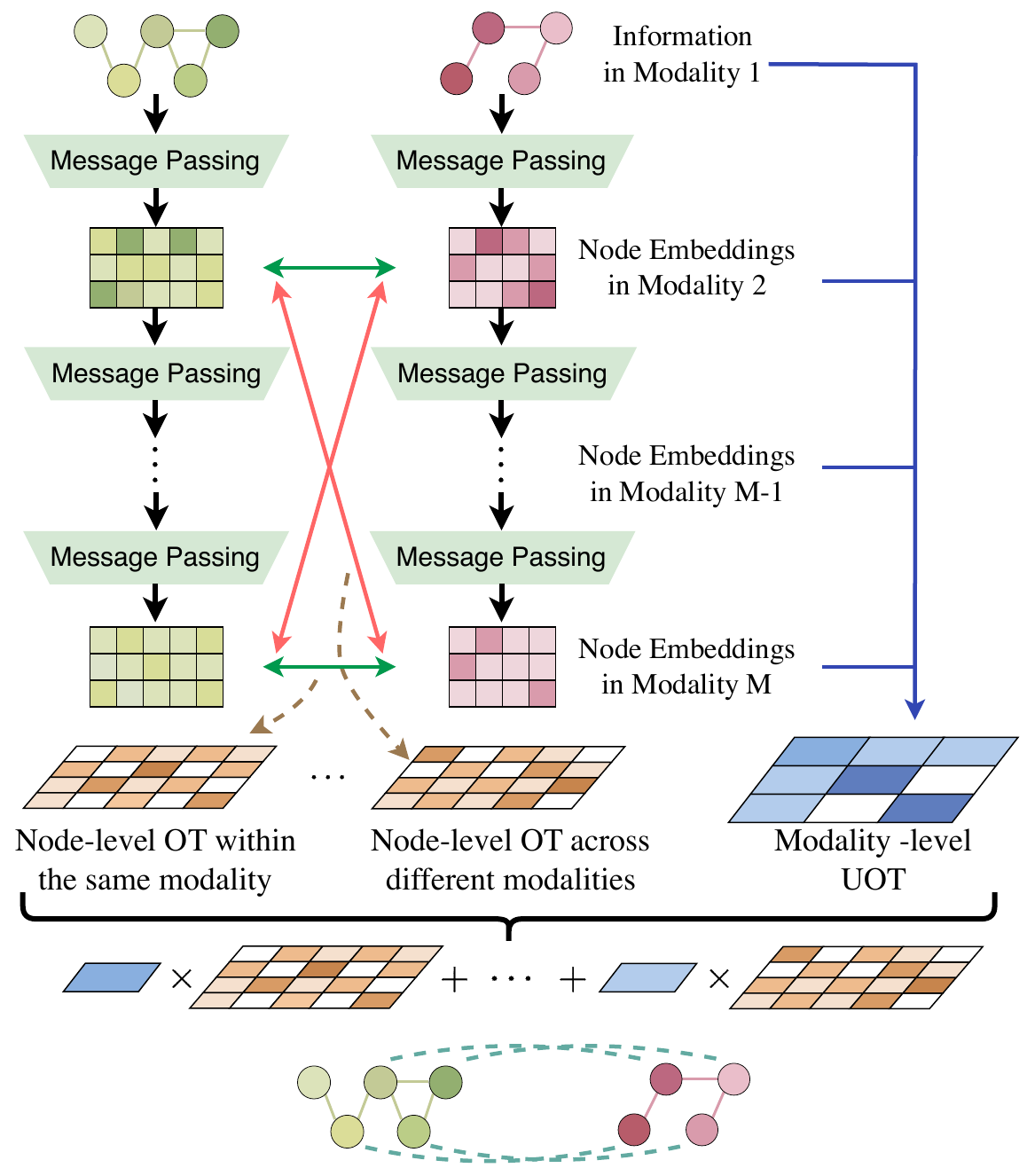}
    \caption{The scheme of our method. 
    Given two graphs, we extract their multi-modal information by multi-layer message passing.
    We align the node embeddings of the two graphs within the same modality and across different modalities, respectively, by solving a series of node-level OT problems. 
    We fuse the alignment results by solving a modality-level UOT problem and infer node correspondence accordingly.}
    \label{fig:scheme}
\end{figure}

In practice, achieving exact graph matching is always challenging because of its NP-hardness. 
Therefore, many methods have been developed to match graphs approximately.
Classic graph matching methods often formulate the task as a quadratic assignment problem (QAP)~\cite{loiola2007survey} based on graphs' adjacency matrices~\cite{umeyama1988eigendecomposition,koutra2013big,zaslavskiy2008path} or other relation matrices~\cite{zhou2015factorized,hashemifar2016joint,zanfir2018deep,xu2019gromov}.
Recently, some learning-based graph matching methods~\cite{heimann2018regal,trung2020adaptive,gao2021unsupervised} embed graph nodes and then align the node embeddings across different graphs. 
However, most existing methods merely apply specific information from a single modality (e.g., adjacency matrices, node attributes, or subgraph structures), leading to non-robust matching performance.
Although some recent methods match graphs based on multi-modal information~\cite{tang2023robust,titouan2019optimal}, they often apply over-simplified mechanisms to fuse the multi-modal information, resulting in sub-optimal performance. 
To our knowledge, few existing graph matching approaches consider fully leveraging the multi-modal information hidden in graphs, let alone study the impacts of the cross-modal information on the matching results.

To overcome the above problems and fill in the blank, in this study we consider the multi-modal information of graphs and their interactions in graph matching tasks, proposing a robust graph matching method based on an unbalanced hierarchical optimal transport (UHOT) framework. 
As illustrated in Figure~\ref{fig:scheme}, our method formulates the graph matching task as an unbalanced hierarchical optimal transport problem. 
Given two graphs, we apply multi-layer message passing to generate their layer-wise node embeddings. 
The node embeddings obtained in each layer correspond to a modality, reflecting the structural information of the graphs at a specific smoothing strength.
For the two graphs, we align their node embeddings within the same modality and across different modalities, respectively. 
Each alignment is achieved by computing the Gromov-Wasserstein (GW) distance~\cite{memoli2011gromov} (or its variant~\cite{titouan2019optimal}) between the corresponding node embedding sets, and the optimal transport (OT) plan associated with the distance indicates a node-level alignment result.
Enumerating all modality pairs, we consider the weighted average of their corresponding OT plans as the graph matching result, in which the weights are learned by solving a modality-level unbalanced optimal transport (UOT) problem. 

Solving the node-level and modality-level OT problems iteratively leads to the proposed UHOT framework, in which the node-level OT plans provide the alignment results based on different modalities' information and the modality-level UOT plan determines the fusion mechanism of the alignment results. 
In the modality level, solving the UOT problem, in which the significance of different modalities is learned with regularization, helps avoid trivial solutions commonly in existing multi-modal graph matching methods~\cite{tang2023robust} and thus improves the robustness of our method.
We consider different implementations of the UHOT framework, including applying different OT distances~\cite{memoli2011gromov,titouan2019optimal} and selecting different optimization algorithms~\cite{cuturi2013sinkhorn} for the OT problems, and discuss the complexity and application scenarios of the implementations.

\textbf{Different from existing graph matching methods, the proposed UHOT-based method, to our knowledge, first leverages the node alignment results across different modalities in an explicit way and demonstrates their contributions to improving final matching performance.}
It provides a new technical route seldom considered before for robust graph matching.
We test our method in both synthetic and real-world graph matching tasks and compare it with state-of-the-art unsupervised and semi-supervised graph matching methods.
Comprehensive experiments demonstrate the superiority of our method and its robustness. 

\section{Related Work}
Optimal transport (OT) distance and its variants (like GW and FGW distances) provide an effective metric for probability measures (e.g., distributions).
In particular, the OT distance in the Kantorovich form is called Wasserstein distance~\cite{kantorovich1942translocation}, which corresponds to computing an optimal transport plan between two probability measures.
Given the samples of two probability measures, the optimal transport plan between them is formulated as a doubly stochastic matrix indicating the pairwise coherency of the samples~\cite{su2017order,memoli2011gromov}. 
Because of this excellent property, OT distance has received great attention in extensive matching tasks, such as shape matching~\cite{solomon2016entropic}, generative modeling~\cite{genevay2018learning}, and image-text alignment~\cite{chen2020graph}.
In graph analysis, OT distance is also gradually being adopted for graph-to-graph comparisons.
Based on the GW distance, a series of OT-based graph matching methods have been proposed and achieved encouraging performance.
GWL~\cite{xu2019gromov} is the first GW-based method that jointly learns the node embeddings and finds the node correspondence between two graphs.
The FGW distance in~\cite{titouan2019optimal} extends GW distance by considering the Wasserstein term for node attributes, so that it can be applied to match attribute graphs.
SLOTAlign~\cite{tang2023robust} combines GW distance with multi-view structure learning to enhance graph representation power and reduce the effect of structure and feature inconsistency inherited across graphs.

Recently, hierarchical optimal transport (HOT)~\cite{schmitzer2013hierarchical,alvarez2018structured}, as a generalization of original OT, is proposed to compare the distributions with structural information, e.g., measuring the distance between different Gaussian mixture models~\cite{chen2018optimal}.
By solving OT plans at different levels, HOT has achieved encouraging performance in multi-modal distribution matching~\cite{lee2019hierarchical,luo2022differentiable}, multi-modal learning~\cite{luo2022differentiable}, and neural architecture search~\cite{yang2023hotnas}.
To our knowledge, however, these HOT techniques have not yet been attempted in graph matching tasks. 
Additionally, unlike existing HOT work, our UHOT method leverages unbalanced optimal transport (UOT) at the modality level. 
As demonstrated in~\cite{frogner2015learning,chizat2018scaling}, compared to solving classic OT problems, solving UOT problems helps improve the robustness of domain adaptation~\cite{fatras2021unbalanced} and generative modeling~\cite{YangU19}.

\section{Proposed Method}
\subsection{Preliminaries and Motivation}
In this study, we denote a graph as $\mathcal{G} = (\mathcal{V}, \bm{A}, \bm{X})$.
Here, $\mathcal{V}$ is the set of nodes. 
$\bm{A} \in \{0, 1\}^{|\mathcal{V}| \times |\mathcal{V}|}$ is the adjacency matrix, where $A_{ij} = 1$ denotes the presence of an edge between nodes $i$ and $j$, and $A_{ij} = 0$ indicates the absence of an edge.
$\bm{X}=[\bm{x}] \in \mathbb{R}^{|\mathcal{V}| \times d}$ denotes the node attribute matrix, where $|\mathcal{V}|$ represents the number of nodes, and each node has an attribute vector $\bm{x}\in\mathbb{R}^d$. 
Given two graphs, i.e., $\mathcal{G}_s=(\mathcal{V}_s, \bm{A}_s, \bm{X}_s)$ and $\mathcal{G}_t=(\mathcal{V}_t, \bm{A}_t, \bm{X}_t)$, graph matching aims to find the correspondence between their nodes.
The node correspondence can be formulated as a matrix $\bm{T}^*=[T^*_{ij}] \in \mathbb{R}^{|\mathcal{V}_s| \times |\mathcal{V}_t|}$: for each $i\in\mathcal{V}_s$, we can infer its correspondence in $\mathcal{V}_t$ by $j^*=\arg\max_{j\in\mathcal{V}_t} T^*_{ij}$.
Without the loss of generality, in the following content, we assume that $|\mathcal{V}_s|\leq |\mathcal{V}_t|$.

As aforementioned, classic graph matching methods often formulate the task as a QAP problem~\cite{loiola2007survey}, i.e.,
\begin{eqnarray}
\begin{aligned}
    \sideset{}{_{\bm{T}\in\mathcal{P}_{|\mathcal{V}_s|\times|\mathcal{V}_t|}}}\max\langle\bm{D}_s\bm{T}\bm{D}_t,~\bm{T}\rangle.
\end{aligned}
\end{eqnarray}
where the correspondence matrix $\bm{T}$ is formulated as a permutation matrix, and its feasible domain is denoted as $\mathcal{P}_{|\mathcal{V}_s|\times|\mathcal{V}_t|}=\{\bm{T}\in\{0,1\}^{|\mathcal{V}_s|\times|\mathcal{V}_t|}|\bm{T}\bm{1}_{|\mathcal{V}_t|}=\bm{1}_{|\mathcal{V}_s|},\bm{T}^{\top}\bm{1}_{|\mathcal{V}_s|}\leq \bm{1}_{|\mathcal{V}_t|}\}$.
$\bm{D}_s\in\mathbb{R}^{|\mathcal{V}_s|\times|\mathcal{V}_s|}$ and $\bm{D}_t\in\mathbb{R}^{|\mathcal{V}_t|\times|\mathcal{V}_t|}$ are two relation matrices capturing the structural information of the two graphs, respectively. 
In practice, the relation matrices can be implemented as the adjacency matrices~\cite{umeyama1988eigendecomposition,koutra2013big,zaslavskiy2008path} (i.e., $\bm{D}_s=\bm{A}_s$ and $\bm{D}_t=\bm{A}_t$), the node similarity matrices~\cite{zhou2015factorized,caetano2009learning,zanfir2018deep} (i.e., $\bm{D}_s=\bm{X}_s\bm{X}_s^{\top}$ and $\bm{D}_t=\bm{X}_t\bm{X}_t^{\top}$), or their fusion results~\cite{hashemifar2016joint,xu2019gromov}. 

When relaxing the correspondence matrix to a doubly stochastic matrix, i.e., $\bm{T}\in\Omega(\bm{\mu}_s,\bm{\mu}_t)=\{\bm{T}\geq\bm{0}|\bm{T}\bm{1}_{|\mathcal{V}_t|}=\bm{\mu}_s,\bm{T}^{\top}\bm{1}_{|\mathcal{V}_s|}=\bm{\mu}_t\}$, where $\bm{\mu}_s$ and $\bm{\mu}_t$ are two predefined node distributions that indicate the significance of nodes, we can reformulate the above QAP problem as computing a Gromov-Wasserstein (GW) distance between two graphs~\cite{memoli2011gromov}, i.e.,
\begin{eqnarray}\label{eq:gwd}
    \begin{aligned}
        &d_{GW}(\mathcal{G}_s, \mathcal{G}_t)\\ 
        := &\sideset{}{_{\bm{T} \in \Omega(\bm{\mu}_s, \bm{\mu}_t)}}\min \sideset{}{_{i,j,k,l}}\sum |D^s_{ij}-D^t_{kl}|^2 T_{ik}T_{jl} \\
        = &\sideset{}{_{\bm{T} \in \Omega(\bm{\mu}_s, \bm{\mu}_t)}}\min \mathbb{E}_{i,k,j,l\sim\bm{T}\times \bm{T}}[|D^s_{ij}-D^t_{kl}|^2],
    \end{aligned}
\end{eqnarray}
where $D_{ij}^s$ is the element of $\bm{D}_s$ corresponding to the node pair $(i,j)$ in $\mathcal{G}_s$, and similarly, $D_{kl}^t$ is the element of $\bm{D}_t$ corresponding to the node pair $(k,l)$ in $\mathcal{G}_t$.
The GW distance provides a valid distance metric for the collections of graphs~\cite{chowdhury2019gromov}.
In statistics, it computes the minimum expectation of the discrepancy of node pairs (i.e., the $|D^s_{ij}-D^t_{kl}|^2$ in~\eqref{eq:gwd}). 
The doubly stochastic matrix corresponding to the minimum expectation, denoted as $\bm{T}^*$, is called the optimal transport (OT) plan, which can be viewed as a joint distribution of the nodes between the two graphs.
Accordingly, the element in $\bm{T}^*$ indicates the correspondence of the graphs' nodes.
Compared with the original QAP problem, the GW distance is much easier to compute~\cite{titouan2019optimal,xu2019gromov}, making it a promising graph matching method.

The relation matrices, which contain the structural information of graphs, are crucial for the matching performance. 
Constructing the relation matrices purely based on a single modality (e.g., adjacency matrices or node attributes) often leads to non-robust matching results because the structural information of a single modality is sensitive to data noise~\cite{gao2021unsupervised,tang2023robust,xu2019scalable,koutra2013big}.
To overcome this robustness issue, some attempts have been made to leverage multi-modal information in graph matching tasks.
Typically, the work in~\cite{titouan2019optimal} proposes a variant of GW distance, called fused Gromov-Wasserstein (FGW) distance, considering the optimal transport based on both relation matrices and node attributes, i.e.,
\begin{eqnarray}\label{eq:fgwd}
\begin{aligned}
    &d_{FGW}(\mathcal{G}_s, \mathcal{G}_t;\beta)\\
        := &\sideset{}{_{\bm{T} \in \Omega(\bm{\mu}_s, \bm{\mu}_t)}}\min 
        (1-\beta)\underbrace{\sideset{}{_{i,k}}\sum \|\bm{x}_i^s-\bm{x}_k^t\|_2^2 T_{ik}}_{\mathbb{E}_{i,k\sim\bm{T}}[\|\bm{x}_i^s-\bm{x}_k^t\|_2^2]}\\
        &\quad\quad\quad\quad+\beta\underbrace{\sideset{}{_{i,j,k,l}}\sum |D^s_{ij}-D^t_{kl}|^2 T_{ik}T_{jl}}_{\mathbb{E}_{i,k,j,l\sim\bm{T}\times \bm{T}}[|D^s_{ij}-D^t_{kl}|^2]},
\end{aligned}
\end{eqnarray}
where the first term is the Wasserstein term computing the expectation of the distance for node attribute pairs, and the second term is the GW term corresponding to the expectation in~\eqref{eq:gwd}. 
The FGW distance aims to find the OT plan minimizing these two terms jointly, in which the hyperparameter $\beta\in [0,1]$ controls their significance. 
When $\beta=1$, the FGW distance degrades to the GW distance in~\eqref{eq:gwd}, which matches two graphs based on a pair of relation matrices.
Similarly, when $\beta=0$, the FGW distance degrades to the Wasserstein distance~\cite{villani2009optimal} between node attributes. 

Besides the FGW-based matching method, the SLOTAlign in~\cite{tang2023robust} constructs the $\bm{D}_s$ and $\bm{D}_t$ in~\eqref{eq:gwd} by fusing multi-modal relation matrices linearly, i.e., 
\begin{eqnarray}\label{eq:fuse}
    \bm{D}_s=\sideset{}{_{m=1}^{M}}\sum\alpha_m\bm{D}_s^{(m)},\quad\bm{D}_t=\sideset{}{_{m=1}^{M}}\sum \alpha_m\bm{D}_t^{(m)}. 
\end{eqnarray}
Here, $M$ is the number of modalities, and $\{\bm{D}_s^{(m)},\bm{D}_t^{(m)}\}$ is the relation matrix pair corresponding to the $m$-th modality. 
$\bm{\alpha}=[\alpha_m]\in\Delta^{M-1}$ is a learnable parameter vector defined in $(M-1)$-Simplex, which determines the significance of the modalities.
Typically, the relation matrices in different modalities are constructed based on different information, e.g., node attributes, adjacency matrices, and various graph kernels~\cite{borgwardt2005shortest,shervashidze2009efficient,shervashidze2011weisfeiler}.
Given the $\bm{D}_s$ and $\bm{D}_t$ in~\eqref{eq:fuse}, SLOTAlign matches graphs by computing their GW distance.

The above methods have demonstrated that multi-modal information indeed helps improve the robustness of graph matching.
However, their over-simplified linear fusion mechanisms limit the utilization of the multi-modal information.
In particular, the linear fusion step itself eliminates the identifiability of different modalities.\footnote{For example, merely based on the fused matrix $\bm{D}_s$ in~\eqref{eq:fuse}, we cannot obtain its multi-modal components $\{\bm{D}_s^{(m)}\}_{m=1}^{M}$.} 
As a result, none of the existing methods consider the potential of matching graphs across different modalities, which may result in sub-optimal performance. 
In the following content, we propose a robust graph matching method using an unbalanced hierarchical optimal transport framework, which provides a new paradigm to leverage multi-modal information in graph matching tasks.

\subsection{Proposed UHOT Framework}

\subsubsection{Multi-modal Information Extraction}
\begin{figure}[t]
    \centering
    \includegraphics[width=0.95\linewidth]{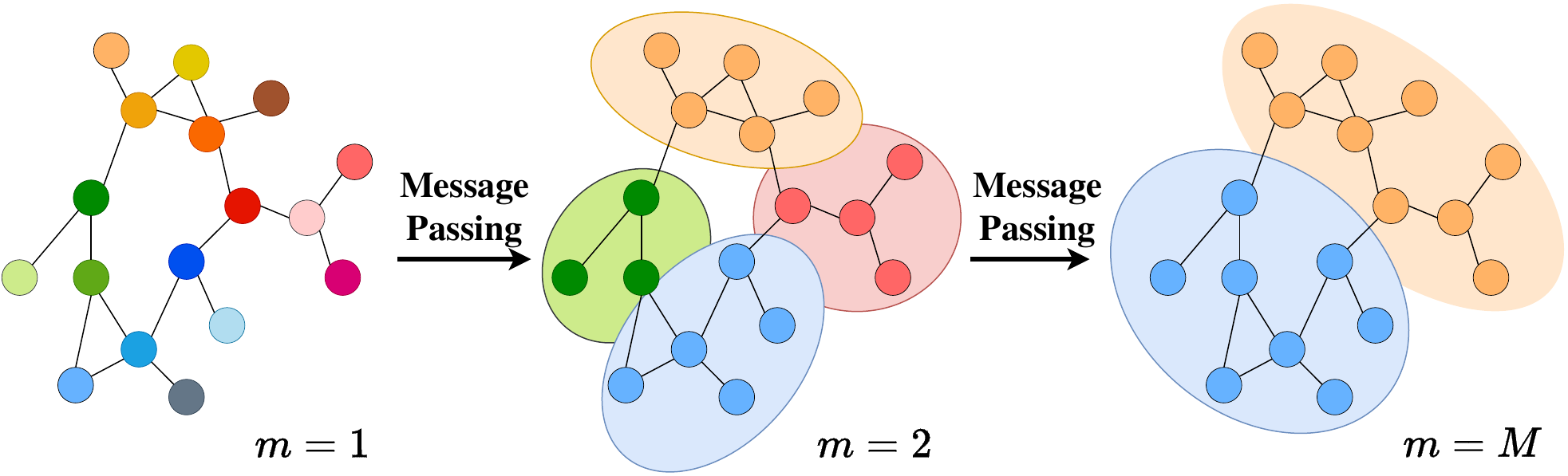}
    \caption{An illustration of our message passing-based multi-modal information extraction.}
    \label{fig:mp}
\end{figure}

In this study, we apply a set of non-learnable message passing layers to extract $M$ modalities' information hidden in a graph.
Typically, given a graph $\mathcal{G}(\mathcal{V},\bm{A},\bm{X})$, we treat the initial node attribute matrix $\bm{X}$ as the information of the first modality. 
The information of the $m$-th modality is derived by passing $\bm{X}$ through $m-1$ message passing layers as follows
\begin{eqnarray}\label{eq:mp}
    \bm{X}^{(m)}=\widehat{\bm{A}}\bm{X}^{(m-1)}=\widehat{\bm{A}}^{m-1}\bm{X},\quad m=1,...,M,
\end{eqnarray} 
where $\widehat{\bm{A}} = \bm{M}^{-\frac{1}{2}}(\bm{A} + \bm{I})\bm{M}^{-\frac{1}{2}}$ is the symmetric normalized adjacency matrix with self-loop, $\bm{I}$ is the identity matrix, and $\bm{M}$ is the degree matrix of $\bm{A} + \bm{I}$.
From the perspective of graph spectral filtering, each message passing layer in~\eqref{eq:mp} (i.e., $\widehat{\bm{A}}\bm{X}^{(m-1)}$) works as a low-pass filter of the current node embeddings.
With the increase in the number of message passing layers, the smoothness of the node embeddings increases accordingly. 
As a result, the node embeddings derived by different layers encode the structural information of the graph (e.g., the node clustering structure) in different granularity levels, as illustrated in Figure~\ref{fig:mp}. 

Denote the node embeddings of the $M$ modalities as a set $\mathcal{X}=\{\bm{X}^{(m)}\}_{m=1}^{M}$.
Inspired by SLOTAlign~\cite{tang2023robust}, we can further define a set of relational matrices for the graph, i.e., $\mathcal{D}=\{\bm{D}^{(m)}\}_{m=1}^{M}$, where
\begin{eqnarray}\label{eq:relation_mat}
\begin{aligned}
    &\bm{D}^{(1)}=\bm{A},\\
    &\bm{D}^{(m)}=\bm{X}^{(m-1)}(\bm{X}^{(m-1)})^{\top},~m=2,...,M.
\end{aligned}
\end{eqnarray}
We can reformulate a graph based on the multi-modal information, denoted as $\mathcal{G}=\{\mathcal{G}^{(m)}(\mathcal{V},\bm{D}^{(m)},\bm{X}^{(m)})\}_{m=1}^{M}$, where each $\mathcal{G}^{(m)}(\mathcal{V},\bm{D}^{(m)},\bm{X}^{(m)})$ encodes the graph structural information of the $m$-th modality.

\subsubsection{Node-level Optimal Transports across Different Modalities}
Given two graphs with multi-modal information, denoted as $\mathcal{G}_s=\{\mathcal{G}_s^{(m)}(\mathcal{V}_s,\bm{D}_s^{(m)},\bm{X}_s^{(m)})\}_{m=1}^{M}$ and $\mathcal{G}_t=\{\mathcal{G}_t^{(m)}(\mathcal{V}_t,\bm{D}_t^{(m)},\bm{X}_t^{(m)})\}_{m=1}^{M}$, respectively, we can align their nodes by computing their node-level optimal transport plans within each modality and across different modalities, respectively.
In particular, we can construct a distance matrix with size $M\times M$, i.e., $\bm{D}(\mathcal{G}_s,\mathcal{G}_t;\beta)=[d_{FGW}(\mathcal{G}_s^{(p)},\mathcal{G}_t^{(q)};\beta)]$, where $d_{FGW}(\mathcal{G}_s^{(p)},\mathcal{G}_t^{(q)};\beta)$ is the FGW distance between the graphs in the $p$-th and $q$-th modalities, respectively. 
We compute each $d_{FGW}(\mathcal{G}_s^{(p)},\mathcal{G}_t^{(q)};\beta)$ by solving~\eqref{eq:fgwd}. 
An associated optimal transport plan is derived as 
\begin{eqnarray}
\begin{aligned}
    &\bm{T}^{(p,q)}\\
=&\arg\sideset{}{_{\bm{T}\in\Pi(\bm{\mu}_s,\bm{\mu}_t)}}\min (1-\beta)\mathbb{E}_{i,k\sim\bm{T}}[\|\bm{x}_i^{s,(p)}-\bm{x}_k^{t,(q)}\|_2^2]\\
&\quad\quad\quad\quad\quad\quad+\beta\mathbb{E}_{i,k,j,l\sim\bm{T}\times \bm{T}}[|D^{s,(p)}_{ij}-D^{t,(q)}_{kl}|^2]
\end{aligned}
\end{eqnarray} 
Following existing work~\cite{xu2019gromov,xu2020gromov}, we set the node distributions $\bm{\mu}_s$ and $\bm{\mu}_t$ to be uniform.

\begin{itemize}
\item \textbf{Remark 1.} 
The OT plan $\bm{T}^{(p,q)}$ indicates the node-level alignment results of the two graphs based on the $p$-th and $q$-th modalities, respectively.  
When $p=q$, $\bm{T}^{(p, q)}$ captures the node correspondence between $\mathcal{G}_s$ and $\mathcal{G}_t$ within the same modality. 
When $p\neq q$, $\bm{T}^{(p, q)}$ captures the node correspondence across different modalities.
Different from existing methods, we enumerate all modality pairs and compute $M^2$ OT plans explicitly, i.e., $\mathcal{T} = \{ \bm{T}^{(p, q)} \}_{p=1, q=1}^{M}$, which explicitly considers the cross-modal alignment results.
\end{itemize}

\subsubsection{A Modality-level Unbalanced Optimal Transport}
For the two graphs, their modality pairs generally contribute to their matching with different significance. 
Therefore, given $\mathcal{T} = \{ \bm{T}^{(p, q)} \}_{p=1, q=1}^{M}$, we need to determine the weight of each $\bm{T}^{(p,q)}$ automatically. 
In this study, we achieve this aim by solving an unbalanced optimal transport problem at the modality level.
Specifically, taking the distance matrix $\bm{D}(\mathcal{G}_s,\mathcal{G}_t;\beta)$ as the grounding cost, we compute the minimum Wasserstein distance between two learnable modalities' distributions, i.e.,
\begin{eqnarray}\label{eq:wd}
\begin{aligned}
    &\sideset{}{_{\bm{\nu}_s,\bm{\nu}_t\in\Delta^{M-1}}}\min d_W(\bm{\nu}_s, \bm{\nu}_t; \bm{D}(\mathcal{G}_s,\mathcal{G}_t;\beta))\\ 
    =&\sideset{}{_{\bm{\nu}_s,\bm{\nu}_t,\bm{\Theta}}}\min \sideset{}{_{p,q=1}^M}\sum \theta_{pq} d_{FGW}(\mathcal{G}_s^{(p)},\mathcal{G}_t^{(q)};\beta)\\
    =&\sideset{}{_{\bm{\nu}_s,\bm{\nu}_t,\bm{\Theta}}}\min\langle\bm{D}_{}, \bm{\Theta}\rangle\\
    &s.t.~\bm{\nu}_s,\bm{\nu}_t\in\Delta^{M-1},~~\bm{\Theta}\in \Omega(\bm{\nu}_s, \bm{\nu}_t)
\end{aligned}
\end{eqnarray}
Here, $\bm{\nu}_s,\bm{\nu}_t\in\Delta^{M-1}$ are two learnable vectors in the $(M-1)$-Simplex, indicating the significance of the $M$ modalities for $\mathcal{G}_s$ and $\mathcal{G}_t$, respectively. 
$\Omega(\bm{\nu}_s, \bm{\nu}_t)$ is the set of the doubly-stochastic matrices that take $\bm{\nu}_s$ and $\bm{\nu}_t$ as marginals. 
$\bm{\Theta} = [\theta_{pq}]\in \Omega(\bm{\nu}_s, \bm{\nu}_t)$ is the transport matrix defined for the modalities.
It can be explained as a joint distribution of the modalities corresponding to different graphs, and its element $\theta_{pq}$ represents the coherency probability of the $p$-th modality of $\mathcal{G}_s$ and the $q$-th modality of $\mathcal{G}_t$.
\begin{itemize}
    \item \textbf{Remark 2.} In principle, the matrix $\bm{\Theta}$ indicates the significance of different modality pairs.
    When the coherency probability of the modality pair $(p,q)$ (i.e., $\theta_{pq}$) is large, the corresponding distance $d_{FGW}(\mathcal{G}_s^{(p)},\mathcal{G}_t^{(q)};\beta)$ should be small, which means that $\mathcal{G}_s^{(p)}$ and $\mathcal{G}_t^{(q)}$ are matched well and their matching result $\bm{T}^{(p,q)}$ is significant. 
\end{itemize}
Note that, because $\bm{\nu}_s$ and $\bm{\nu}_t$ are learnable, the optimal solution of~\eqref{eq:wd} may set them as one-hot vectors, so that only the $\theta_{pq}$ associated with the minimum $d_{FGW}(\mathcal{G}_s^{(p)},\mathcal{G}_t^{(q)};\beta)$ is one, while the remaining $\theta$'s are zeros.
To avoid such a trivial solution, we further introduce a regularizer for $\bm{\nu}_s$ and $\bm{\nu}_t$, penalizing their KL-divergence to the uniform distribution $\frac{1}{M}\bm{1}_M$, i.e.,
\begin{eqnarray}\label{eq:kld}
\begin{aligned}
    R(\bm{\nu}_s,\bm{\nu}_t)=KL\Bigl(\bm{\nu}_s\|\frac{1}{M}\bm{1}_M\Bigr) +  KL\Bigl(\bm{\nu}_t\|\frac{1}{M}\bm{1}_M\Bigr)
\end{aligned}
\end{eqnarray}
Plugging~\eqref{eq:kld} into~\eqref{eq:wd} leads to the well-known unbalanced optimal transport (UOT) problem~\cite{frogner2015learning,chizat2018scaling}. 

\subsubsection{Robust Graph Matching by Minimizing HOT} 
The composition of the above two-level optimal transport problems leads to a HOT distance between the graphs, i.e.,
\begin{eqnarray}\label{eq:hot}
\begin{aligned}
    d_{HOT}(\mathcal{G}_s, \mathcal{G}_t) := d_{W}(\bm{\nu}_s, \bm{\nu}_t; \bm{D}(\mathcal{G}_s,\mathcal{G}_t;\beta)),
\end{aligned}
\end{eqnarray}
where the grounding cost $\bm{D}$ is constructed by the node-level FGW distances with the hyperparameter $\beta$, and the Wasserstein distance computes the modality-level optimal transport plan.
Taking the regularizer in~\eqref{eq:kld} into account, we can match two graphs by computing an unbalanced hierarchical optimal transport (UHOT) distance between them, i.e., 
\begin{eqnarray}\label{eq:UHOT}
\begin{aligned}
    &\mathcal{T},\bm{\Theta},\bm{\nu}_t,\bm{\nu}_s\\ =&\arg\underbrace{\sideset{}{_{\mathcal{T},\bm{\Theta},\bm{\nu}_t,\bm{\nu}_s}}\min d_W(\bm{\nu}_s, \bm{\nu}_t; \bm{D}_{})
    + R(\bm{\nu}_s,\bm{\nu}_t)}_{d_{UHOT}(\mathcal{G}_s,\mathcal{G}_t)}.
\end{aligned}
\end{eqnarray}
This problem corresponds to the computation of the $M^2$ node-level optimal transport plans $\mathcal{T}=\{\bm{T}^{(p,q)}\}_{p,q=1}^M$ and the unbalanced modality-level optimal transport plan $\bm{\Theta}$.
We call this optimization problem ``UHOT'' because the marginals $\bm{\nu}_s$ and $\bm{\nu}_t$ are learnable variables regularized by the KL-divergence terms.

Given optimized $\mathcal{T}$ and $\bm{\Theta}$, we compute the final matching result as the weighted sum of all $\bm{T}^{(p,q)}$'s, i.e., 
\begin{eqnarray}
    \label{eq: final result}
    \begin{aligned}
        \bm{T} = \sideset{}{_{p,q=1}^{M}}\sum \theta_{pq} \bm{T}^{(p,q)}.
    \end{aligned}
\end{eqnarray}
Accordingly, the final matching result is dominated by the $\bm{T}^{(p,q)}$'s corresponding to the significant modality pairs.

\section{Optimization Algorithm}
\label{sec:alg}
\begin{figure}[t]
\centering
\subfigure[$d_{GW}(\mathcal{G}_s^{(1)}, \mathcal{G}_t^{(1)})$]{\includegraphics[height=4.0cm]{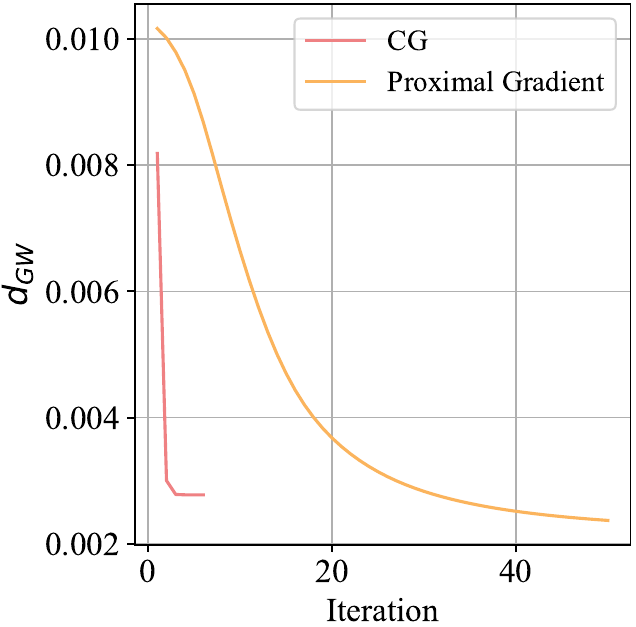}\label{fig:curve_gw}
}
\subfigure[$d_{FGW}(\mathcal{G}_s^{(1)}, \mathcal{G}_t^{(1)})$]{
\includegraphics[height=4.0cm]{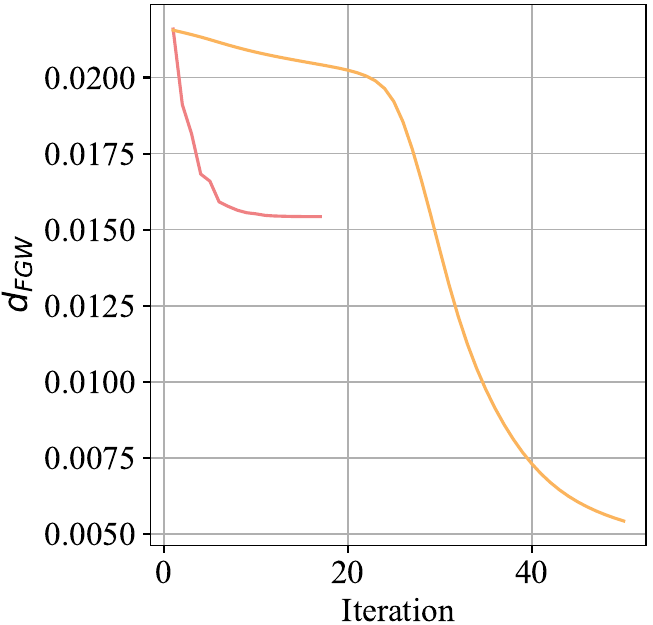}\label{fig:curve_fgw}
}
\caption{The convergence curve on PPI.}
\label{fig:curve}
\end{figure}

We propose a bi-level learning algorithm to solve the UHOT problem in~\eqref{eq:UHOT}. 
In this study, we apply the proximal gradient algorithm~\cite{xu2019scalable} or the conditional gradient (CG) algorithm~\cite{titouan2019optimal} to compute each FGW distance efficiently.
We first reformulate the FGW distance between two graphs as follows.
\begin{eqnarray}\label{eq:fgwd_appendix}
\begin{aligned}
    &d_{FGW}(\mathcal{G}_s, \mathcal{G}_t;\beta)\\
    :=& \sideset{}{_{\bm{T} \in \Omega(\bm{\mu}_s, \bm{\mu}_t)}}\min (1-\beta)\sideset{}{_{i,k}}\sum \|\bm{x}_i^s-\bm{x}_k^t\|_2^2 T_{ik}\\
    &\quad\quad\quad\quad+ \beta\sideset{}{_{i,j,k,l}}\sum |D^s_{ij}-D^t_{kl}|^2 T_{ik}T_{jl}\\
    =& \sideset{}{_{\bm{T} \in \Omega(\bm{\mu}_s, \bm{\mu}_t)}}\min \langle (1-\beta)\bm{X}_s \bm{X}_t^{\top} + \beta\bm{L}(\bm{D}_s, \bm{D}_t, \bm{T}), \bm{T} \rangle,
\end{aligned}
\end{eqnarray}
where $\bm{L}(\bm{D}_s, \bm{D}_t, \bm{T}) = (\bm{D}_s \odot \bm{D}_s)\bm{\mu}_s \bm{1}_{|\mathcal{V}_t|}^{\top} + \bm{1}_{|\mathcal{V}_s|}\bm{\mu}_t^{\top}(\bm{D}_t \odot \bm{D}_t)^{\top} - 2\bm{D}_s \bm{T} \bm{D}_t^{\top}$ and $\odot$ denotes the Hadamard product of matrix.  
$\bm{X}_s$ and $\bm{X}_t$ are two node attribute matrices.
The proximal gradient algorithm~\cite{xu2019scalable} decomposes a complicated non-convex optimization problem into a series of convex sub-problems. 
The global convergence of this proximal gradient method is guaranteed in~\cite{xu2019gromov}.
Algorithm~\ref{alg: PPA} gives the pipeline of the proximal gradient algorithm.
The conditional gradient algorithm~\cite{titouan2019optimal} introduces a linear regularization term, where the solution provides a descent direction and a line-search whose optimal step can be found in closed form to update the FGW distance.
Algorithm~\ref{alg: CG} gives the pipeline of the conditional gradient algorithm.
Figure~\ref{fig:curve} shows the convergence of these two algorithms in computing the GW and FGW distances. 
It can be observed that the CG algorithm converges faster, but the proximal gradient converges to smaller values.


\begin{algorithm}[t]
    \caption{The proximal gradient algorithm for computing $d_{FGW}(\mathcal{G}_s,\mathcal{G}_t;\beta)$} \label{alg: PPA}
	\begin{algorithmic}[1]
	    \REQUIRE $\mathcal{G}_s(\mathcal{V}_s,\bm{D}_s,\bm{X}_s),\mathcal{G}_t(\mathcal{V}_t,\bm{D}_t,\bm{X}_t)$, trade-off parameter $\beta$, marginals $\bm{\mu}_s$, $\bm{\mu}_t$, matching matrix $\bm{T}$, entropic regularizer $\lambda$, the number of outer/inner iterations $\{M,N\}$.
            \STATE Initialize $\bm{b} = \bm{\mu}_t$ and $\bm{T}^{(0)}=\bm{T}$
            \STATE $\bm{K}^{(0)} = (1-\beta)\bm{X}_s \bm{X}_t^{\top} + \beta\bm{L}(\bm{D}_s, \bm{D}_t, \bm{T}^{(0)})$
            \FOR{$m = 1, \cdots, M$}
            \STATE $\bm{G} = \exp{(-\bm{K}^{(m-1)} / \lambda)} \odot \bm{T}^{(m-1)}$
            \FOR{$n = 1, \cdots, N$}
            \STATE $\bm{a} = \bm{\mu}_s / (\bm{G}\bm{b} )$, and then $\bm{b} = \bm{\mu}_t / (\bm{G}^\top\bm{a} )$
            \ENDFOR
            \STATE $\bm{T}^{(m)}= \operatorname{diag}(\bm{a}) \bm{G} \operatorname{diag}(\bm{b})$
            \ENDFOR
            \STATE $d_{FGW}(\mathcal{G}_s,\mathcal{G}_t;\beta) = \langle \bm{K}^{(M)}, \bm{T}^{(M)}\rangle$
            \RETURN $d_{FGW}(\mathcal{G}_s,\mathcal{G}_t;\beta)$, and $\bm{T}\leftarrow \bm{T}^{(M)}$.
	\end{algorithmic}
\end{algorithm}

\begin{algorithm}[t]
    \caption{The conditional gradient (CG) algorithm for computing $d_{FGW}(\mathcal{G}_s,\mathcal{G}_t;\beta)$} \label{alg: CG}
	\begin{algorithmic}[1]
	    \REQUIRE $\mathcal{G}_s(\mathcal{V}_s,\bm{D}_s,\bm{X}_s),\mathcal{G}_t(\mathcal{V}_t,\bm{D}_t,\bm{X}_t)$, trade-off parameter $\beta$, marginals $\bm{\mu}_s$, $\bm{\mu}_t$, matching matrix $\bm{T}$, the number of iterations $M$.
            \STATE Initialize $\bm{T}^{(0)}=\bm{T}$
            \STATE $\bm{K}^{(0)} = (1-\beta)\bm{X}_s\bm{X}_t^{\top} + \beta\bm{L}(\bm{D}_s, \bm{D}_t, \bm{T}^{(0)})$
            \FOR{$m = 1, \cdots, M$}
            \STATE $\bm{G} = (1-\beta)\bm{X}_s \bm{X}_t^{\top} + 2\beta\bm{L}(\bm{D}_s, \bm{D}_t, \bm{T}^{(m-1)})$ 
            \STATE $\tilde{\bm{T}}^{(m)} = \arg\min_{\bm{T} \in \Omega(\bm{\mu}_s, \bm{\mu}_t)} \langle \bm{G}, \bm{T} \rangle$ 
            \STATE Line-search to check whether $d_{FGW}(\mathcal{G}_s,\mathcal{G}_t;\beta)$ decreases given $\tilde{\bm{T}}^{(m)}$ and determine a momentum $\tau \in (0, 1)$
            \STATE $\bm{T}^{(m)} = (1 - \tau) \bm{T}^{(m-1)} + \tau \tilde{\bm{T}}^{(m)}$
            \ENDFOR
           \STATE $d_{FGW}(\mathcal{G}_s,\mathcal{G}_t;\beta) = \langle \bm{K}^{(M)}, \bm{T}^{(M)}\rangle$
            \RETURN $d_{FGW}(\mathcal{G}_s,\mathcal{G}_t;\beta)$, and $\bm{T} \leftarrow \bm{T}^{(M)}$.
	\end{algorithmic}
\end{algorithm}

\begin{algorithm}[t]
    \caption{Compute $d_W(\bm{\nu}_s, \bm{\nu}_t; \bm{D}(\mathcal{G}_s,\mathcal{G}_t;\beta))$}\label{alg:marginal}
	\begin{algorithmic}[1]
	    \REQUIRE Marginals $\bm{\nu}_s$, $\bm{\nu}_t$, cost matrix $\bm{D}_{}$, the number of modalities $M$, entropic regularizer $\lambda$, learning rate $\gamma$, the number of Sinkhorn-scaling iterations $N$.
            \STATE $\bm{b}^{(0)} = \frac{1}{M}\bm{1}_{M}$, $\bm{\Theta} = \bm{\nu}_s\bm{\nu}_t^\top$, and $\bm{Q} = \exp{(-\bm{D}_{}/ \lambda)}$
	    \FOR{$n=1,...,N$}
	        \STATE $\bm{a}^{(n)} = \bm{\nu}_s / (\bm{Q} \bm{b}^{(n-1)})$, $ \bm{b}^{(n)} = \bm{\nu}_t / (\bm{Q}^\top \bm{a}^{(n)})$
	    \ENDFOR
         \STATE $\bm{\nu}_s = \bm{\nu}_s - \gamma(\log\bm{a} - \frac{\log\bm{a}^\top\bm{1}}{\bm{Q}}\bm{1}) / \lambda$
         \STATE $\bm{\nu}_t = \bm{\nu}_t - \gamma(\log\bm{b} - \frac{\log\bm{b}^\top\bm{1}}{\bm{Q}}\bm{1}) / \lambda$
          \RETURN $\bm{\Theta} \leftarrow \operatorname{diag}(\bm{a}^{(N)}) \bm{Q} \operatorname{diag}(\bm{b}^{(N)})$, $\bm{\nu}_s$, and $\bm{\nu}_t$.
	\end{algorithmic}
\end{algorithm}

In theory, both of the algorithms ensure that the variables converge to a stationary point~\cite{xu2019gromov,lacoste2016convergence}.
Typically, for a graph with $V$ nodes and $M$ modalities, the computational complexity of the algorithms is $\mathcal{O}(M^2V^3)$.
Fortunately, because the inner product of node embeddings constructs the relation matrices we applied, we can reduce the complexity of the algorithms to $\mathcal{O}(M^2V^2d)$ by leveraging the low-rank structures of the relation matrices~\cite{scetbon2022linear}. 

In the modality-level, we need to $i)$ compute the OT plan $\bm{\Theta}$ associated with the Wasserstein distance and $ii)$ update the marginals $\bm{\nu}_s$ and $\bm{\nu}_t$. 
We achieve these two steps jointly in a Sinkhorn-based algorithmic framework.
Specifically, we rewrite~\eqref{eq:wd} by introducing an entropic regularizer, i.e., 
\begin{eqnarray}\label{eq:eot}
    \begin{aligned}
    \sideset{}{_{\bm{\nu}_s,\bm{\nu}_t\in\Delta^{M-1},\bm{\Theta}\in \Omega(\bm{\nu}_s, \bm{\nu}_t)}}\min\langle\bm{D}_{}, \bm{\Theta}\rangle + \lambda H(\bm{\Theta}),
    \end{aligned}
\end{eqnarray}
where $H(\bm{\Theta}) = \langle\bm{\Theta}, \log \bm{\Theta} \rangle$, and $\lambda$ is the hyperparameter controlling the importance of the entropy term.
This entropic regularizer improves the smoothness of the original problem.
Following existing work~\cite{cuturi2013sinkhorn,xie2021a,luo2022differentiable}, the entropic OT problem in~\eqref{eq:eot} can be solved efficiently by the Sinkhorn-scaling algorithm, whose computational complexity is $\mathcal{O}(M^2)$.
When updating the marginals, we apply the gradient descent algorithm~\cite{frogner2015learning}, i.e., 
\begin{eqnarray}\label{eq: gradient descent}
    \begin{aligned}
        \bm{\nu}_s \leftarrow \bm{\nu}_s - \gamma \frac{\partial d_w(\bm{\nu}_s, \bm{\nu}_t)}{\partial \bm{\nu}_s},~~
        \bm{\nu}_t \leftarrow \bm{\nu}_t - \gamma \frac{\partial d_w(\bm{\nu}_s, \bm{\nu}_t)}{\partial \bm{\nu}_t},
    \end{aligned}
\end{eqnarray}
where $\gamma$ is the learning rate.
The gradients can be computed efficiently when applying the Sinkhorn-scaling iterations. 
Algorithm~\ref{alg:marginal} shows the corresponding pipeline.

\begin{algorithm}[t]
    \caption{UHOT-based Graph Matching}\label{alg:scheme}
	\begin{algorithmic}[1]
	    \REQUIRE Graphs $\mathcal{G}_s=(\mathcal{V}_s, \bm{A}_s, \bm{X}_s)$ and $\mathcal{G}_t=(\mathcal{V}_t, \bm{A}_t, \bm{X}_t)$, the number of modalities $M$, the number of training iterations $T$.
            \STATE Based on~\eqref{eq:mp} and~\eqref{eq:relation_mat}, obtain $\mathcal{G}_s=\{\mathcal{G}^{(m)}(\mathcal{V}_s,\bm{D}_s^{(m)},\bm{X}_s^{(m)})\}_{m=1}^{M}$ and $\mathcal{G}_t=\{\mathcal{G}_t^{(m)}(\mathcal{V}_t,\bm{D}_t^{(m)},\bm{X}_t^{(m)})\}_{m=1}^{M}$. 
            \STATE Set $\bm{\mu}_s = \frac{1}{|\mathcal{V}_s|}\bm{1}_{|\mathcal{V}_s|}$, $\bm{\mu}_t = \frac{1}{|\mathcal{V}_t|}\bm{1}_{|\mathcal{V}_t|}$
            \STATE Initialize $\bm{\nu}_s = \frac{1}{M}\bm{1}_{M}$, $\bm{\nu}_t = \frac{1}{M}\bm{1}_{M}$, $\bm{T} = \bm{\mu}_s \bm{\mu}_t^\top$
            \FOR{$t=1,\cdots,T$}
    	    \FOR{$p,q=1,\cdots,M$}
                \STATE Get $d_{FGW}(\mathcal{G}_s^{(p)},\mathcal{G}_t^{(q)};\beta)$ and $\bm{T}^{(p,q)}$ via the proximal gradient algorithm~\cite{xu2019scalable} or the conditional gradient algorithm~\cite{titouan2019optimal}.
    	    \ENDFOR
         \STATE Get $\bm{D}(\mathcal{G}_s,\mathcal{G}_t;\beta)=[d_{FGW}(\mathcal{G}_s^{(p)},\mathcal{G}_t^{(q)};\beta)]$, $\mathcal{T} = \{ \bm{T}^{(p, q)} \}$.
          \STATE Optimize $\Theta$ via Sinkhorn-scaling algorithm~\cite{cuturi2013sinkhorn}, and update $\bm{\nu}_s$, $\bm{\nu}_t$ by the gradient descent in~\cite{frogner2015learning}. 
          \STATE $\bm{T} = \sum_{p=1,q=1}^{M}\theta_{pq} \bm{T}^{(p,q)}$.
          \ENDFOR
	\end{algorithmic}
\end{algorithm}

In summary, our method first computes $M^2$ distances and derives the corresponding optimal transport plans, each of which indicates a matching result for graph nodes.
Then, the significance of the $M$ modalities is computed by solving an entropic OT problem with learnable marginals.
The final matching result is obtained by aggregating all the matching plans according to~\eqref{eq: final result}.
Algorithm~\ref{alg:scheme} gives the pipeline of our method.
The computational complexity of Algorithm~\ref{alg:scheme} is $O(T(M^2V^2d + M^2N))$, where $T$ is the number of outer loops and $N$ is the number of Sinkhorn-scaling iterations.

\section{Advantages Compared to Existing Methods}
Our UHOT-based method provides a generalized framework for OT-based graph matching, and many existing methods can be viewed as its simplified special cases.
\subsubsection{Compared to single-modal graph matching methods}
As aforementioned, solving~\eqref{eq:UHOT} without the regularizer leads to the following trivial solution
\begin{eqnarray}\label{eq:trivial1}
\begin{aligned}
    &\sideset{}{_{\mathcal{T},\bm{\Theta},\bm{\nu}_t,\bm{\nu}_s}}\min d_W(\bm{\nu}_s, \bm{\nu}_t; \bm{D})\\
    \Leftrightarrow &\sideset{}{_{p,q\in\{1,..,M\}}}\min d_{FGW}(\mathcal{G}_s^{(p)},\mathcal{G}_t^{(q)};\beta),
\end{aligned}
\end{eqnarray}
in which the optimal $\bm{\nu}_t$ and $\bm{\nu}_s$ are one-hot vectors, and their non-zero elements indicate the modality pair $(p,q)$ that corresponds to the minimum FGW distance. 
When $p=q$, this trivial solution corresponds to matching $\mathcal{G}_s$ and $\mathcal{G}_t$ based on a single modality.
When $p=q=1$, only the original node attributes and adjacency matrices are applied to compute the FGW distance, and our UHOT-based method degrades to the single-modal strategy in~\cite{titouan2019optimal}.
When further setting $\beta=1$, the FGW distance is specified as the GW distance, and our UHOT-based method is equivalent to the GWL method in~\cite{xu2019gromov,xu2019scalable}. 
Introducing the regularizer in~\eqref{eq:kld} allows us to effectively leverage the cross-modal alignment results (i.e., $\{\bm{T}^{(p,q)}\}_{p\neq q}$). 

\subsubsection{Compared to multi-modal graph matching methods}
As a representative multi-modal graph matching method, SLOTAlign~\cite{tang2023robust} obtains an OT plan $\bm{T}^*$ shared by all modality pairs by computing $d_{GW}(\sum_{m=1}^{M}\alpha_m\bm{D}_{s}^{(m)},\sum_{m=1}^{M}\alpha_m\bm{D}_{t}^{(m)})$. 
It is easy to find that SLOTAlign can be treated as a special case of our UHOT-based method --- when setting $\beta=1$, the solution of SLOTAlign is also a feasible (non-optimal) solution of~\eqref{eq:UHOT}, i.e., $\bm{\nu}_s=\bm{\nu}_t=\bm{\alpha}$, $\bm{\Theta}=\text{diag}(\bm{\alpha})$, and $\mathcal{T}=\{\bm{T}^{(p,q)}=\bm{T}^*\}_{p,q=1}^{K}$. 
In other words, SLOTAlign only considers the node-level alignment results within the same modality, and only the GW distance between each modality's relation matrices is involved.
Because of introducing the alignment results across different modalities and leveraging FGW distance, our UHOT-based method can be more robust than SLOTAlign.
Additionally, although introducing a proximal gradient algorithm to update $\bm{\alpha}$, SLOTAlign cannot prevent $\bm{\alpha}$ from being one-hot vector in theory because it does not consider the regularization of $\bm{\alpha}$.

As another special case of our method, we can set $\bm{\nu}_s=\bm{\nu}_t=\frac{1}{M}\bm{1}_M$, and the UHOT distance between graphs degrades to the classic HOT distance, i.e., $d_W(\frac{1}{M}\bm{1}_M,\frac{1}{M}\bm{1}_M;\bm{D}(\mathcal{G}_s,\mathcal{G}_t;\beta))$. 
Furthermore, we have the following proposition: 
\begin{proposition}
For simplifying notations, we define $\bm{D}_s^{all}:=\sum_{p=1}^M\bm{D}_s^{(p)}$ and $\bm{D}_t^{all}:=\sum_{q=1}^M\bm{D}_t^{(q)}$, respectively. 
When setting $\beta=1$ (using GW distance as the grounding cost), we have
\begin{eqnarray}\label{eq:trivial2}
    \begin{aligned}
    &d_W\Bigl(\frac{1}{M}\bm{1}_M,\frac{1}{M}\bm{1}_M;\bm{D}(\mathcal{G}_s,\mathcal{G}_t;1)\Bigr)\\  
    \leq &\frac{1}{M^2}(d_{GW}(\bm{D}_s^{all}, \bm{D}_t^{all})+C).
    \end{aligned}
\end{eqnarray}
where $C$ is nonnegative and defined as
\begin{eqnarray*}
\begin{aligned}
    \sideset{}{_{p,q=1}^{M}}\sum\operatorname{tr}( (\bm{D}_s^{(p)} \odot \bm{D}_s^{(p)})\bm{\mu}_s \bm{\mu}_s^{\top} ) +\operatorname{tr}(\bm{\mu}_t \bm{\mu}_t^{\top} (\bm{D}_t^{(q)} \odot \bm{D}_t^{(q)})^{\top})\\
    -\operatorname{tr}( (\bm{D}_s^{all} \odot \bm{D}_s^{all})\bm{\mu}_s \bm{\mu}_s^{\top} ) -\operatorname{tr}(\bm{\mu}_t \bm{\mu}_t^{\top} (\bm{D}_t^{all} \odot \bm{D}_t^{all})^{\top}).
\end{aligned}
\end{eqnarray*}
\end{proposition}
\begin{proof}

Denote $\bm{T}^*$ as the optimal solution of $d_{GW}(\bm{D}_s^{all}, \bm{D}_t^{all})$, i.e.,
\begin{eqnarray}
    \begin{aligned}
    \bm{T}^* &= \arg\sideset{}{_{\bm{T} \in \Omega(\bm{\mu}_s, \bm{\mu}_t)}}\min d_{GW}( \bm{D}_s^{all}, \bm{D}_t^{all} )\\
    	&= \arg\sideset{}{_{\bm{T} \in \Omega(\bm{\mu}_s, \bm{\mu}_t)}}\min \langle \bm{L}( \bm{D}_s^{all}, \bm{D}_t^{all}, \bm{T} ), \bm{T} \rangle.
    \end{aligned}
\end{eqnarray}
For the matrix $\bm{L}$, we denote its component $(\bm{D}_s \odot \bm{D}_s)\bm{\mu}_s \bm{1}_{|\mathcal{V}_t|}^{\top} + \bm{1}_{|\mathcal{V}_s|}\bm{\mu}_t^{\top}(\bm{D}_t \odot \bm{D}_t)^{\top}$ as $\bm{C}(\bm{D}_s, \bm{D}_t)$. 
Since $\bm{T} \in \Omega(\bm{\mu}_s, \bm{\mu}_t)$, we have
\begin{eqnarray}\label{eq:ind}
    \begin{aligned}
        &\langle\bm{C}(\bm{D}_s, \bm{D}_t), \bm{T}\rangle\\
        =&\operatorname{tr}( (\bm{D}_s \odot \bm{D}_s)\bm{\mu}_s \bm{\mu}_s^{\top} ) +\operatorname{tr}(\bm{\mu}_t \bm{\mu}_t^{\top} (\bm{D}_t \odot \bm{D}_t)^{\top} ),
    \end{aligned}
\end{eqnarray}
which indicates that the result of $\langle \bm{C}(\bm{D}_s, \bm{D}_t) , \bm{T}\rangle$ is a constant independent of $\bm{T}$.

In general, the optimal $\bm{\Theta}^*\neq[\frac{1}{M^2}]$ (i.e., a uniform distribution), so we have
\begin{eqnarray}
    \begin{aligned}
    &d_W\Big(\frac{1}{M}\bm{1}_M,\frac{1}{M}\bm{1}_M;\bm{D}(\mathcal{G}_s,\mathcal{G}_t;1)\Big) \\
    \leq &\frac{1}{M^2}\sideset{}{_{p,q=1}^M}\sum d_{GW}(\bm{D}_s^{(p)}, \bm{D}_t^{(q)}).
    \end{aligned}
\end{eqnarray}
Then we have
\begin{eqnarray}\label{eq:upper}
    \begin{aligned}
    &\sideset{}{_{p,q=1}^M}\sum d_{GW}(\bm{D}_s^{(p)}, \bm{D}_t^{(q)}) \\
    \le&\sideset{}{_{p,q=1}^M}\sum \langle \bm{L}( \bm{D}_s^{(p)}, \bm{D}_t^{(q)}, \bm{T}^* ), \bm{T}^* \rangle \\
    =&\sideset{}{_{p,q=1}^M}\sum \langle \bm{C}(\bm{D}_s^{(p)}, \bm{D}_t^{(q)}) - 2\bm{D}_s^{(p)}\bm{T}^* \bm{D}_t^{(q)}, \bm{T}^* \rangle \\
    =&\sideset{}{_{p,q=1}^M}\sum \langle \bm{C}(\bm{D}_s^{(p)}, \bm{D}_t^{(q)}), \bm{T}^* \rangle- 2\langle \bm{D}_s^{all} \bm{T}^* \bm{D}_t^{all}, \bm{T}^* \rangle \\
    =&\sideset{}{_{p,q=1}^M}\sum \langle \bm{C}(\bm{D}_s^{(p)}, \bm{D}_t^{(q)}), \bm{T}^* \rangle + d_{GW}(\bm{D}_s^{all}, \bm{D}_t^{all})\\
     &- \langle \bm{C}(\bm{D}_s^{all}, \bm{D}_t^{all}), \bm{T}^* \rangle \\
    =&\underbrace{\Big\langle \sideset{}{_{p,q=1}^M}\sum\bm{C}(\bm{D}_s^{(p)}, \bm{D}_t^{(q)}) - \bm{C}(\bm{D}_s^{all}, \bm{D}_t^{all}), \bm{T}^*  \Big\rangle}_{\geq 0} \\
    &+ d_{GW}(\bm{D}_s^{all}, \bm{D}_t^{all}).
    \end{aligned}
\end{eqnarray}
where the first term is independent with $\bm{T}^*$ because of~\eqref{eq:ind}.
It is nonnegative because of the Cauchy–Schwarz inequality.
\end{proof}

In other words, when $\beta=1$, such a simplified UHOT distance is comparable to SLOTAlign, given the scaling coefficient $\frac{1}{M^2}$.

\section{Experiments}
\subsection{Experimental Setup}

Denote our UHOT-based graph matching method as \textbf{UHOT-GM}. 
We demonstrate its effectiveness by comparing it with state-of-the-art graph matching methods.
Additionally, we provide comprehensive analytic experiments, verifying the rationality of using cross-modal alignment results and demonstrating the robustness of our method to data noise and hyperparameter settings.
All the experiments are implemented in PyTorch and conducted on an NVIDIA 3090 GPU.
\textbf{Representative results are shown below. 
More implementation details and results are in Appendix.}

\begin{table}[t]
\centering
\caption{Description of the datasets.}
\label{tab:data}
\small{
\begin{tabular}{c|ccc}
\hline
Dataset                                  & \#Nodes & \#Edges  & Dim. of Attr. \\ \hline
\multirow{2}{*}{ACM-DBLP}              & 9,872 & 39,561 & 17    \\
                                       & 9,916 & 44,808 & 17    \\ \hline
\multirow{2}{*}{Douban Online-Offline} & 3,906 & 16,328 & 538   \\
                                       & 1,128 & 3,022  & 538   \\ \hline
Cora                                   & 2,708 & 5,278  & 1,433  \\
                                       \hline
PPI                                    & 1,767 & 17,042  & 50  \\                                         \hline
\end{tabular}
}
\end{table}

\begin{table*}[t!]
\centering
    \caption{Comparison on node correctness (\%). 
    For each dataset, we bold the best three results and highlight the best one in red.}
    \label{tab:cmp}
    \tabcolsep=6pt
    \subtable[Node correctness on real-world datasets]{
    \small{\begin{tabular}{c|c|ccc|ccc}
    \hline
    \multirow{2}{*}{Type}            &\multirow{2}{*}{Method} &  \multicolumn{3}{c|}{ACM-DBLP}                              & \multicolumn{3}{c}{Douban Online-Offline}                 \\ \cline{3-8} 
                            &                                  & NC@1           & NC@5           & NC@10          & NC@1           & NC@5           & NC@10          \\ \hline
    \multirow{3}{*}{semi-}                 & IsoRank & 17.09          & 35.42          & 47.11          & 30.86          & 50.09          & 61.27          \\
    \multirow{3}{*}{supervised}                   & FINAL                                 & 30.25          & 55.32          & 67.95          & 52.24          & \textcolor{red}{\textbf{89.80}} & \textcolor{red}{\textbf{95.97}} \\
    & DeepLink  & 12.19          & 32.98          & 44.58          & 8.86           & 22.36          & 30.95          \\
    & CENALP              & 34.81          & 51.86          & 62.23           & 23.70          & 38.10          & 43.56          \\ \hline
    \multirow{9}{*}{unsupervised}                & UniAlign    & 0.08           & 0.41           & 0.91           & 0.63           & 3.49           & 8.23           \\
                       & REGAL                                 & 3.49           & 9.74           & 13.61          & 1.97           & 6.44           & 10.11          \\
                      & GAlign                                 & 58.43          & 78.78          & 84.46          & 44.10          & 67.98          & 77.73          \\
                      & WAlign                                 & 63.91          & 83.86          & \textbf{89.12}          & 39.53          & 61.63          & 71.02          \\
                    & GWL                                 & 4.02           & 5.96           & 7.34           & 0.27           & 0.72           & 1.07           \\
                    & FGW                                 & 49.11          & 52.06          & 52.09          & \textbf{58.86} & 63.23          & 63.69           \\
                  & SLOTAlign ($M=4$)                                & 65.52          & 84.05          & 87.76         & 49.91          & \textbf{74.69}          & \textbf{79.43}            \\
                \cline{2-8}
              & UHOT-GM ($M=2$)                            & \textbf{67.65}    & \textbf{85.26}    & 88.52   & 54.03        & 67.71        & 70.93   \\ 
              & UHOT-GM ($M=3$)                                & \textbf{69.53} & \textbf{86.97} & \textbf{90.26}  & \textcolor{red}{\textbf{62.97}}          & 71.47          & 75.76             \\ 
              & UHOT-GM ($M=4$)                            & \textcolor{red}{\textbf{70.13}}    & \textcolor{red}{\textbf{87.19}}    & \textcolor{red}{\textbf{90.86}}   & \textbf{59.93}        & \textbf{74.06}        & \textbf{77.28}   \\ 
    \hline
    \end{tabular}
    }}
    \subtable[Node correctness on synthetic datasets]{
    \small{\begin{tabular}{c|c|ccc|ccc}
    \hline
    \multirow{2}{*}{Type}            &\multirow{2}{*}{Method} &  \multicolumn{3}{c|}{PPI}                                   & \multicolumn{3}{c}{Cora}                                     \\ \cline{3-8} 
                            &                                  & NC@1           & NC@5           & NC@10          & NC@1           & NC@5           & NC@10          \\ \hline
    \multirow{3}{*}{semi-}                 & IsoRank & 17.71          & 28.64          & 34.75          & 16.88          & 34.12           & 42.95         \\
    \multirow{3}{*}{supervised}                   & FINAL                                 & 38.09          & 52.91          & 55.35    & 67.25          & 81.35           & 85.52      \\
    & DeepLink  & 10.36          & 14.94          & 18.05          & 10.86          & 27.81           & 36.34         \\
    & CENALP              & 28.35          & 41.43          & 47.82          & 76.55          & 86.85           & 88.81          \\ \hline
    \multirow{9}{*}{unsupervised}                & UniAlign    & 0.68           & 2.77           & 4.92           & 0.41           & 1.85            & 3.91          \\
                       & REGAL                                 & 6.68           & 18.11          & 25.69          & 5.50           & 11.11           & 14.73         \\
                      & GAlign                                 & 67.18          & 78.49          & 82.57          & 98.38          & 99.85          & 99.96          \\
                      & WAlign                                 & 64.63          & 73.23          & 76.91          & 93.72          & 96.01          & 96.38          \\
                    & GWL                                 & 11.38          & 13.30          & 16.07          & 0.03           & 0.11           & 0.37           \\
                    & FGW                                 & 83.32          & 83.32          & 83.32          & 99.19          & 99.19           & 99.19          \\
                  & SLOTAlign ($M=4$)                                & 76.63          & 82.06          & 83.76              & 98.86          & 99.89           & 99.89             \\
                \cline{2-8}
              & UHOT-GM ($M=2$)                            & \textbf{86.64}  & \textbf{90.89}   & \textcolor{red}{\textbf{92.30}}  & \textcolor{red}{\textbf{99.45}} & \textcolor{red}{\textbf{100.00}} & \textcolor{red}{\textbf{100.00}}       \\ 
              & UHOT-GM ($M=3$)                                & \textcolor{red}{\textbf{87.10}} & \textcolor{red}{\textbf{91.06}} &  \textbf{92.13} & \textbf{99.41} & \textcolor{red}{\textbf{100.00}} & \textcolor{red}{\textbf{100.00}}     \\ 
              & UHOT-GM ($M=4$)                            & \textbf{83.93}  & \textbf{89.64}  & \textbf{91.17}   & \textcolor{red}{\textbf{99.45}} & \textcolor{red}{\textbf{100.00}} & \textcolor{red}{\textbf{100.00}}      \\ 
    \hline
    \end{tabular}
    }}
\end{table*}

\subsubsection{Datasets}
In this study, we consider four graph datasets. 
Each dataset contains one or two real-world graphs with their topology and attribute information. 
Table~\ref{tab:data} shows the 
statistics of the datasets.
Details for datasets are described below.
\begin{itemize}
    \item \textbf{ACM-DBLP}~\cite{zhang2016final} is a two co-authorship networks dataset for publication information. The ACM network includes 9,916 authors (i.e., nodes) and 44,808 co-authorships (i.e., edges), while the DBLP network includes 9,872 authors and 39,561 co-authorships. 
    Node attributes are composed of the number of papers published by the author in 17 locations. 
    The 6,325 co-authors in the two networks constitute the ground-truth node correspondence.
    \item \textbf{Douban Online-Offline}~\cite{zhong2012comsoc} includes an online graph with 16,328 interactions among 3,906 users and an offline graph with 3,022 interactions among 1,118 users. 
    The user's location represents node attributes. 
    The ground-truth node correspondence is the 1,118 users appearing in both graphs.
    \item \textbf{Cora}~\cite{yang2016revisiting} is a citation network, whose nodes are publications and edges are citation relations. 
    It has 2,708 nodes and 5,278 edges, and each node has 1,433 attributes.
    \item \textbf{PPI}~\cite{zitnik2017predicting} is a protein-protein interaction network. 
    It contains 1,767 nodes with 50 attributes and 17,042 edges.
\end{itemize}
Since Cora and PPI only contain one graph, we generate the other graph by cutting $E$\% edges in the original graph randomly and adding $E$\% random edges accordingly.
Here, $E\in\{10, 20, ..., 60\}$, indicating different noise levels.

\subsubsection{Baselines} 
For each dataset, we select seven unsupervised graph matching methods as baselines.
Among them, UniAlign~\cite{koutra2013big} is based on solving a QAP problem, REGAL~\cite{heimann2018regal}, WAlign~\cite{gao2021unsupervised}, and GAlign~\cite{trung2020adaptive} are based on node embedding alignment, and GWL~\cite{xu2019gromov}, FGW~\cite{titouan2019optimal}, and SLOTAlign~\cite{tang2023robust} are based on computing OT distances. 
Furthermore, to demonstrate the advantages of UHOT-GM as an unsupervised method, we select four semi-supervised baselines for comparison, including IsoRank~\cite{singh2008global}, FINAL~\cite{zhang2016final}, DeepLink~\cite{zhou2018deeplink} and CENALP~\cite{du2019joint}.
These semi-supervised baselines require partial node pairs as training labels. 
We use 10\% of the ground-truth node pairs when implementing these semi-supervised methods.  

\subsubsection{Hyperparameter Setting.}
Our UHOT-GM applies three message passing layers to generate four modalities, leading to a fair comparison with SLOTAlign. 
Additionally, to demonstrate the efficiency of UHOT-GM in using multi-modal information, we also apply UHOT-GM with two or three modalities, respectively.
By default, we apply FGW distance in UHOT-GM and compute it by the proximal gradient algorithm~\cite{xu2019gromov}, with $\beta\in[0.5,0.9]$.
When solving the modality-level UOT problem, we set the weight of the entropic regularizer in~\eqref{eq:eot} as $\lambda=0.01$ and the learning rate of the modality distributions as $\gamma=1.0$. 
The robustness of our method to the hyperparameters is shown in the following analytic experiments.

\subsubsection{Metrics} 
For each method, we evaluate its performance by the commonly used Top-K node correctness (denoted as NC@K).
In particular, given a node of the graph $\mathcal{G}_s$, NC@K takes the most similar $K$ nodes from all possible matching in the graph $\mathcal{G}_t$ as a Top-K list, and finally calculates the percentage of ground-truth matching in the list. 
Note that, since we implement semi-supervised baselines with 10\% of the ground-truth, we take the ground-truth into account in the final results as well so that we can compare them fairly with unsupervised methods. 
For each dataset, we implement each method five times with different random seeds and report its average performance in the five trials.

\subsection{Numerical Comparisons}
\subsubsection{Node Correctness}
Table~\ref{tab:cmp} shows the matching performance of various methods on the four datasets. 
We can find that UHOT-GM achieves the best NC@1 results on all four datasets, which even outperforms those semi-supervised baselines. 
In particular, the performance of some unsupervised methods, like IsoRank, UniAlign, REGAL, and GWL, is unsatisfactory because they merely leverage the graph topological information (i.e., adjacency matrices) to match graphs while ignoring the utilization of other modalities (e.g., node attributes and subgraph structures). 
On the contrary, the methods applying multi-modal information, including UHOT-GM, often achieve encouraging results.
This phenomenon demonstrates the usefulness of multi-modal information in graph matching tasks.

UHOT-GM performs consistently better than others on ACM-DBLP, PPI, and Cora. 
For the most challenging Douban Online-Offline dataset, where there exists a large disparity in the number of nodes, UHOT-GM still performs the best on NC@1 and achieves comparable results on NC@5 and NC@10. 
This result shows that UHOT-GM remains competitive in those graph matching tasks with extremely imbalanced nodes.
Additionally, the most competitive multi-modal baseline, SLOTAlign, applies four modalities, while the UHOT-GM using three modalities can overcome its performance on NC@1.
This phenomenon implies that $i)$ compared to SLOTAlign, UHOT-GM can leverage multi-modal information of graphs more effectively, and $ii)$ taking cross-modal alignment results into account indeed contributes to improved matching performance.

\subsubsection{Robustness to Noise}
\begin{figure*}[t]
\centering
\subfigure[Robustness to noise]{\includegraphics[height=6.0cm]{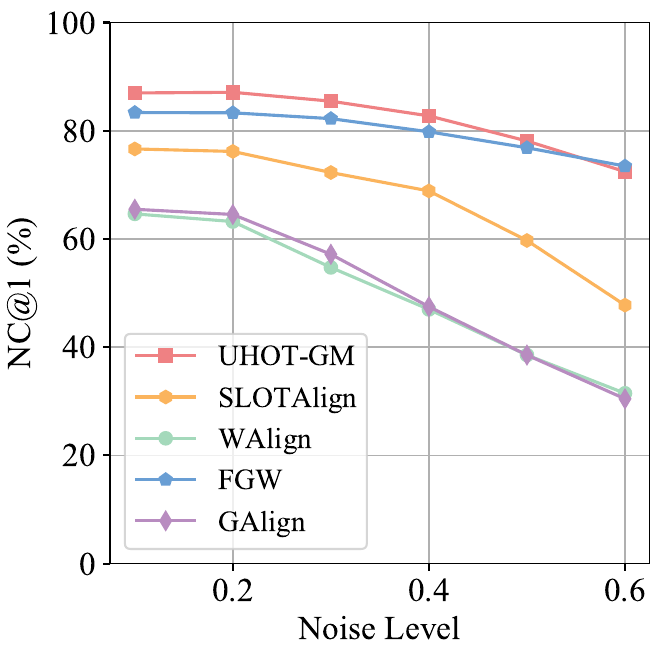}\label{fig:robust}
}
\subfigure[Runtime comparison]{
\includegraphics[height=6.0cm]{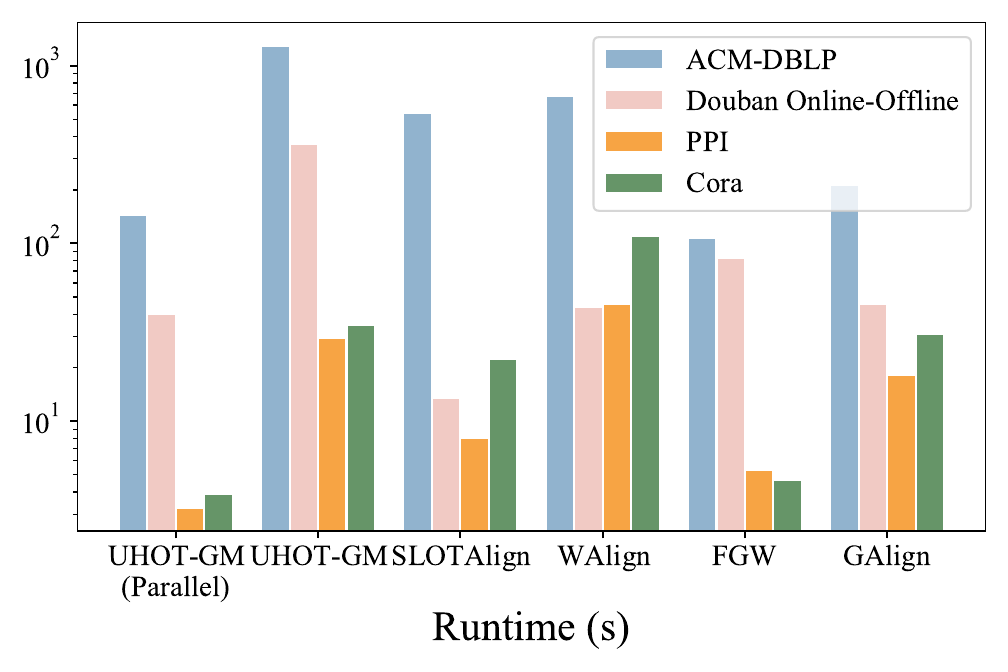}\label{fig:runtime}
}
\caption{Comparisons on robustness and efficiency.}
\label{fig:cmp2}
\end{figure*}

Given the PPI graphs, whose ratio of randomly reconnected edges increases from $5\%$ to $60\%$, we test various unsupervised graph matching methods on their robustness to data noise.
Experimental results in Figure~\ref{fig:robust} show that the methods using FGW distance, e.g., FGW and our UHOT-GM, maintain high node correctness even if 60\% of edges are affected by noise. 
On the contrary, SLOTAlign and WAlign consider the GW and Wasserstein distances between graphs, respectively, whose performance is sensitive to noise.
These results indicate that in highly-noisy matching tasks, applying FGW distance, which computes the OT plan based on both node embeddings and relation matrices, helps improve the robustness of the OT-based matching methods.

\subsubsection{Runtime Comparison}
Figure~\ref{fig:runtime} shows the comparison for various unsupervised graph matching methods on runtime.
We can find that the runtime of UHOT-GM is comparable to that of WAlign and GAlign. 
Compared to FGW and SLOTAlign, UHOT-GM takes longer time in general because it computes multiple FGW distances. 
Taking the improvement in node correctness into account, we think the computational complexity of UHOT-GM is tolerable. 
Moreover, the FGW distances involved in UHOT-GM can be computed in parallel, so the runtime of UHOT-GM in practice can be comparable to that of FGW and SLOTAlign as well, as shown in the ``UHOT-GM (Parallel)'' group of Figure~\ref{fig:runtime}.

\begin{table*}[t!]
\centering
\caption{Ablation study on using different modalities.}
\label{tab:modality}
\small{
\begin{tabular}{c|ccc|ccc}
\hline
Used & \multicolumn{3}{c|}{ACM-DBLP} & \multicolumn{3}{c}{Douban Online-Offline} \\ \cline{2-7} 
Modalities                          & NC@1          & NC@5          & NC@10               & NC@1                & NC@5                & NC@10 \\ \hline
Proposed      & \textbf{70.13}         & \textbf{87.19}         & \textbf{90.86}               & \textbf{59.93}               & \textbf{74.06}               & \textbf{77.28}\\ 
Only Low-pass    & 40.76         & 60.40         & 67.98               & 10.91                & 26.39               & 27.28\\ 
Add High-pass    & 68.57      & 85.82      & 90.03  & 35.51        & 72.81        & 76.74   \\
\hline
\end{tabular}
}
\end{table*}

\subsection{Ablation Study}

\subsubsection{The Rationality of Proposed Message Passing}
The message passing layers used in UHOT-GM work as low-pass graph filters.
They extract graph structural information in different granularity levels. 
In general, these low-pass modalities are insensitive to the noise imposed on graphs.
As a result, UHOT-GM leverages these low-pass modalities jointly with the original graph structural information (i.e., node attributes and adjacency matrices) to achieve graph matching robustly. 
Here, two questions arise: 
$i)$ Can we achieve robust graph matching purely based on the low-pass modalities? 
$ii)$ Can high-pass modalities lead to robust graph matching?
To answer these two questions, we consider two variants of the proposed message passing mechanism. 
In particular, ``Only Low-pass'' means that we only apply the last two layers' embeddings (i.e., the low-pass modalities) as the multi-modal information to match graphs.
``Add High-pass'' means that besides the original modalities, we further take the high-pass graph filtering result, i.e., $\bm{X}_{H}=\widehat{\bm{L}}\bm{X}$, where $\widehat{\bm{L}}$ is normalized graph Laplacian matrix, as an additional modality and match graphs accordingly.

Table~\ref{tab:modality} shows the graph matching results achieved by the UHOT-GM using different message-passing mechanisms.
We can find that the proposed message-passing mechanism achieves the best performance, while the above two variants lead to performance degradation.
Firstly, when only considering the low-pass modalities, we lose the information on original node attributes and adjacency matrices, which harms the matching results. 
Secondly, applying the high-pass modality to graph matching tasks may be inappropriate. 
In particular, graph matching is naturally sensitive to the topological noise (e.g., the random connections and disconnections of edges) in graphs~\cite{trung2020adaptive,tang2023robust}, while the high-pass graph filtering encodes the discrepancy of node attributes along graph edges, whose output is largely influenced by the noise of the edges. 
In summary, the results in Table~\ref{tab:modality} demonstrate the rationality of our method.

\begin{figure}[t]
    \centering
    \subfigure[The impact of $\gamma$]{
    \includegraphics[width=0.95\linewidth]{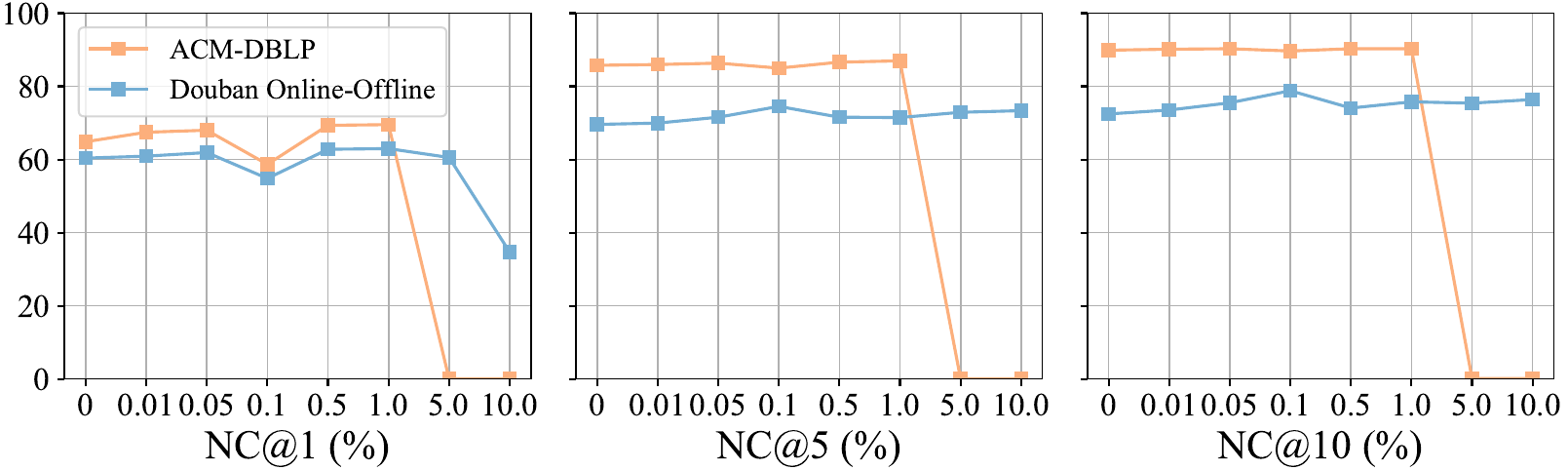}\label{fig:gamma}
    }
    \subfigure[The impact of $\beta$]{
    \includegraphics[width=0.95\linewidth]{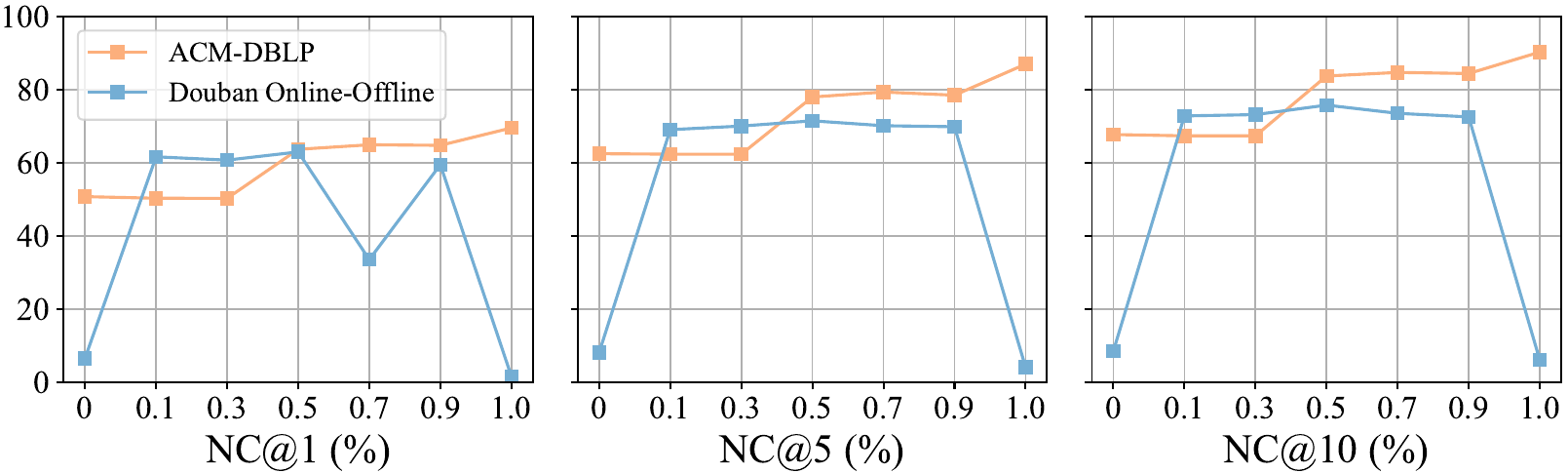}\label{fig:beta}
    }
    \subfigure[The impact of $\lambda$]{
    \includegraphics[width=0.95\linewidth]{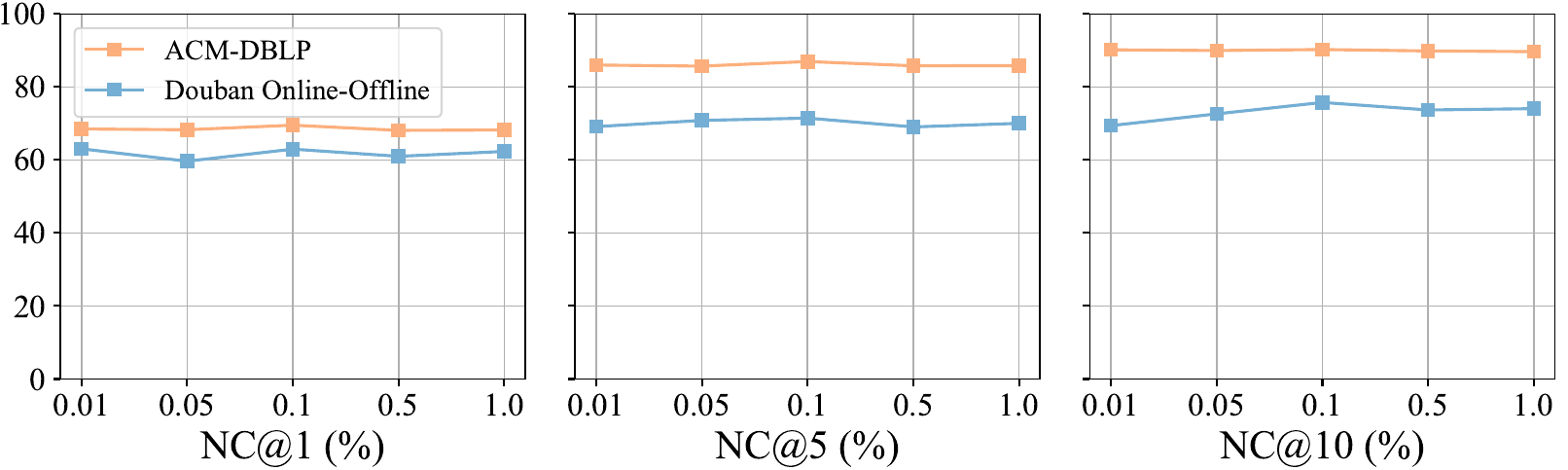}\label{fig:lambda}
    }
    \caption{Testing on the robustness to hyperparameters.}
\end{figure}

\begin{figure}[t]
    \centering
    \subfigure[Illustrations of the modality-level OT plans]{
    \includegraphics[width=0.95\linewidth]{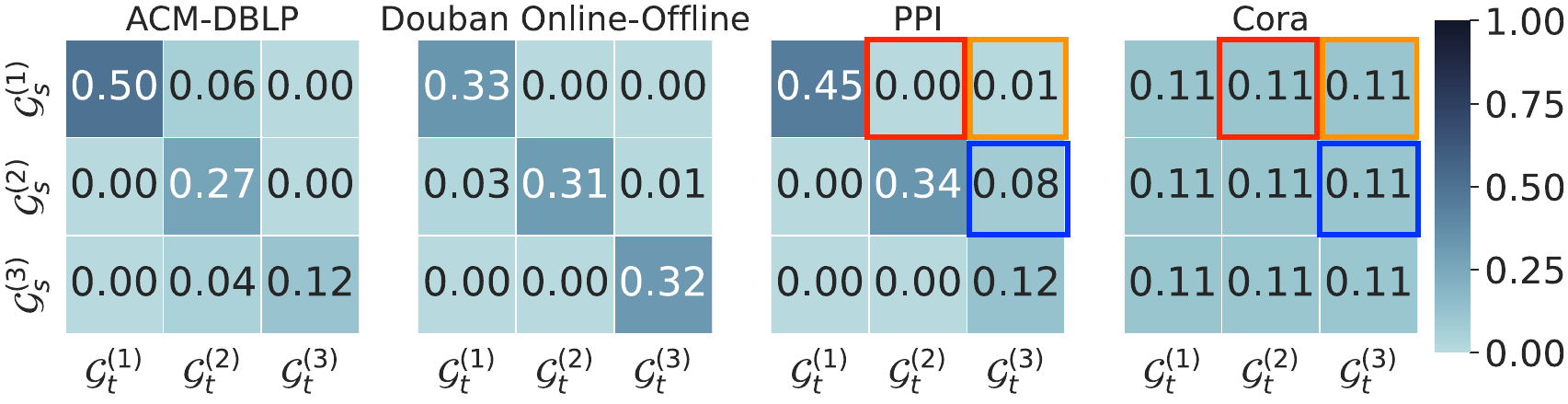}\label{fig:modality_plans}
    }
    \subfigure[Ground truth and some cross-modal alignment results]{
    \includegraphics[width=0.95\linewidth]{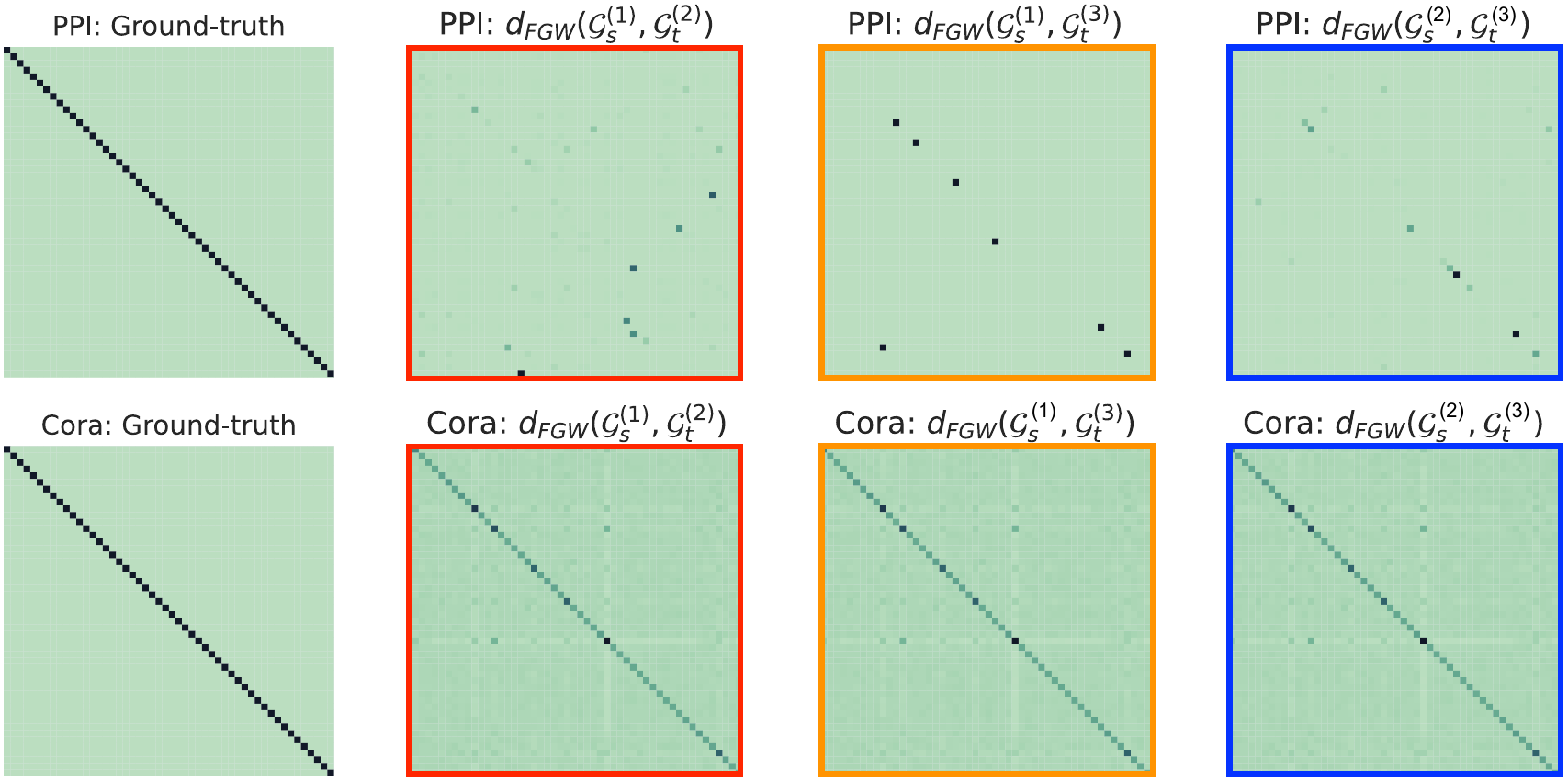}\label{fig:node_plans}
    }
    \caption{(a) The modality-level OT plans, in which some modality pairs are marked.
    (b) The node-level OT plans corresponding to the marked modality pairs. 
    For the convenience of visualization, we take the first 50 nodes for all datasets.}
    \label{fig:cross-modal}
\end{figure}

\subsubsection{The Robustness to Key Hyperparameters}
Our UHOT-GM method has three key hyperparameters, including the learning rate $\gamma$ of modalities' significance, the weight $\beta$ in FGW distance, and the weight $\lambda$ of the KL-divergence regularizer. 
Taking the learning rate $\gamma$ from $\{0.01, 0.05, 0.1, 0.5, 1, 5, 10\}$, we explore its impact on matching results in Figure~\ref{fig:gamma}.
In particular, when $\gamma=0$, it means that we fix $\bm{\nu}_s=\bm{\nu}_t=\frac{1}{M}\bm{1}_M$, treating each modality evenly.
When $\gamma>0$, we update $\bm{\nu}_s$ and $\bm{\nu}_t$ iteratively, with the corresponding learning rate.
The results in Figure~\ref{fig:gamma} show that the UHOT-GM is robust to $\gamma$ in a wide range (i.e., $0.01\leq\gamma\leq 1$), and the best performance is achieved when $\gamma=1$.
When the learning rate is too large, the update of $\bm{\nu}_s$ and $\bm{\nu}_t$ becomes too aggressive and leads to undesired results.
In Figure~\ref{fig:beta}, we explore the impact of $\beta$ on the matching results of two datasets. 
We can find that when $\beta\in [0.1,0.9]$, the NC@5 and NC@10 of UHOT-GM are relatively stable, which demonstrates the robustness of UHOT-GM to $\beta$. 
Based on the results in Figure~\ref{fig:beta}, we can set $\beta\in [0.5, 0.9]$ robustly in practice.
Similarly, UHOT-GM is also robust to the weight $\lambda$ in the range $[0.01, 1]$, as shown in Figure~\ref{fig:lambda}.

\subsubsection{The Rationality of Cross-modal Alignment}
In Figure~\ref{fig:modality_plans}, we visualize the modality-level OT plans obtained by UHOT-GM for different datasets. 
The OT plans of ACM-DBLP and Douban Online-Offline are diagonally-dominant, which means that their matching results are mainly based on the node-level alignment within the same modality. 
However, for PPI and Cora, the contributions of cross-modal alignment results become significant.
For PPI, the upper triangle part of its modality-level OT plan has a significant value.
For Cora, its modality-level OT plan is close to a uniform distribution, which means that the node-level alignment within the same modality and those across different modalities contribute evenly to the final matching results. 

We further mark the upper triangle elements in the modality-level OT plans of PPI and Cora (i.e., the color boxes in Figure~\ref{fig:modality_plans}). 
Each mark corresponds to a modality pair, and we visualize the corresponding node-level OT plans in Figure~\ref{fig:node_plans}. 
We can find that for those insignificant modality pairs (e.g., $(\mathcal{G}_s^{(1)},\mathcal{G}_t^{2})$ and $(\mathcal{G}_s^{(1)},\mathcal{G}_t^{3})$ for PPI), their node-level OT plans are distinguished from the ground truth node correspondence.
On the contrary, for those significant modality pairs (e.g., those for Cora), their node-level OT plans are similar to the ground truth node correspondence.
These phenomena demonstrate the rationality of our method --- UHOT-GM can find useful cross-modal alignment results and assign them large weights when inferring node correspondence.

\section{Conclusion and Future Work}
In this work, we propose a novel UHOT framework for graph matching, leveraging multi-modal information of graphs to achieve robust matching results. 
The proposed UHOT framework makes the first attempt to leverage the cross-modal alignment results explicitly in graph matching tasks, and it avoids trivial solutions by considering the unbalanced modality-level optimal transport. 
Experimental results show that the UHOT-based method achieves encouraging performance in unsupervised graph matching tasks and even outperforms those semi-supervised learning methods.
In summary, our work demonstrates the usefulness of OT-based cross-modal alignment in graph matching tasks, which points out a new technical route seldom considered before. 
In the future, we plan to extend the proposed method, applying it to match more complicated graph structures, e.g., hierarchical graphs and hypergraphs. 
At the same time, we would like to introduce stochastic optimization strategies to improve the efficiency of our algorithm.

\bibliographystyle{IEEEtran}
\bibliography{dhot-gm-arxiv.bib}

\begin{thebibliography}{10}
\providecommand{\url}[1]{#1}
\csname url@samestyle\endcsname
\providecommand{\newblock}{\relax}
\providecommand{\bibinfo}[2]{#2}
\providecommand{\BIBentrySTDinterwordspacing}{\spaceskip=0pt\relax}
\providecommand{\BIBentryALTinterwordstretchfactor}{4}
\providecommand{\BIBentryALTinterwordspacing}{\spaceskip=\fontdimen2\font plus
\BIBentryALTinterwordstretchfactor\fontdimen3\font minus
  \fontdimen4\font\relax}
\providecommand{\BIBforeignlanguage}[2]{{%
\expandafter\ifx\csname l@#1\endcsname\relax
\typeout{** WARNING: IEEEtran.bst: No hyphenation pattern has been}%
\typeout{** loaded for the language `#1'. Using the pattern for}%
\typeout{** the default language instead.}%
\else
\language=\csname l@#1\endcsname
\fi
#2}}
\providecommand{\BIBdecl}{\relax}
\BIBdecl

\bibitem{singh2008global}
R.~Singh, J.~Xu, and B.~Berger, ``Global alignment of multiple protein
  interaction networks with application to functional orthology detection,''
  \emph{Proceedings of the National Academy of Sciences}, no.~35, 2008.

\bibitem{liu2017novel}
Y.~Liu, H.~Ding, D.~Chen, and J.~Xu, ``Novel geometric approach for global
  alignment of ppi networks,'' in \emph{Proceedings of the AAAI Conference on
  Artificial Intelligence}, vol.~31, no.~1, 2017.

\bibitem{li2019partially}
C.~Li, S.~Wang, H.~Wang, Y.~Liang, P.~S. Yu, Z.~Li, and W.~Wang, ``Partially
  shared adversarial learning for semi-supervised multi-platform user identity
  linkage,'' in \emph{Proceedings of the 28th {ACM} International Conference on
  Information and Knowledge Management, {CIKM} 2019, Beijing, China, November
  3-7, 2019}.\hskip 1em plus 0.5em minus 0.4em\relax {ACM}, 2019, pp. 249--258.

\bibitem{li2018distribution}
C.~Li, S.~Wang, P.~S. Yu, L.~Zheng, X.~Zhang, Z.~Li, and Y.~Liang,
  ``Distribution distance minimization for unsupervised user identity
  linkage,'' in \emph{Proceedings of the 27th {ACM} International Conference on
  Information and Knowledge Management, {CIKM} 2018, Torino, Italy, October
  22-26, 2018}.\hskip 1em plus 0.5em minus 0.4em\relax {ACM}, 2018, pp.
  447--456.

\bibitem{huang2022auc}
M.~Huang, Y.~Liu, X.~Ao, K.~Li, J.~Chi, J.~Feng, H.~Yang, and Q.~He,
  ``Auc-oriented graph neural network for fraud detection,'' in
  \emph{Proceedings of the ACM Web Conference 2022}, 2022.

\bibitem{hooi2017graph}
B.~Hooi, K.~Shin, H.~A. Song, A.~Beutel, N.~Shah, and C.~Faloutsos,
  ``Graph-based fraud detection in the face of camouflage,'' \emph{ACM
  Transactions on Knowledge Discovery from Data (TKDD)}, no.~4, 2017.

\bibitem{vento2013graph}
M.~Vento and P.~Foggia, ``Graph matching techniques for computer vision,'' in
  \emph{Image Processing: Concepts, Methodologies, Tools, and Applications},
  2013.

\bibitem{fey2020deep}
M.~Fey, J.~E. Lenssen, C.~Morris, J.~Masci, and N.~M. Kriege, ``Deep graph
  matching consensus,'' in \emph{International Conference on Learning
  Representations}, 2020.

\bibitem{loiola2007survey}
E.~M. Loiola, N.~M.~M. De~Abreu, P.~O. Boaventura-Netto, P.~Hahn, and
  T.~Querido, ``A survey for the quadratic assignment problem,'' \emph{European
  journal of operational research}, no.~2, 2007.

\bibitem{umeyama1988eigendecomposition}
S.~Umeyama, ``An eigendecomposition approach to weighted graph matching
  problems,'' \emph{IEEE transactions on pattern analysis and machine
  intelligence}, no.~5, 1988.

\bibitem{koutra2013big}
D.~Koutra, H.~Tong, and D.~Lubensky, ``Big-align: Fast bipartite graph
  alignment,'' in \emph{2013 IEEE 13th international conference on data
  mining}.\hskip 1em plus 0.5em minus 0.4em\relax IEEE, 2013.

\bibitem{zaslavskiy2008path}
M.~Zaslavskiy, F.~Bach, and J.-P. Vert, ``A path following algorithm for the
  graph matching problem,'' \emph{IEEE Transactions on Pattern Analysis and
  Machine Intelligence}, no.~12, 2008.

\bibitem{zhou2015factorized}
F.~Zhou and F.~De~la Torre, ``Factorized graph matching,'' \emph{IEEE
  Transactions on Pattern Analysis and Machine Intelligence}, vol.~38, no.~9,
  pp. 1774--1789, 2015.

\bibitem{hashemifar2016joint}
S.~Hashemifar, Q.~Huang, and J.~Xu, ``Joint alignment of multiple
  protein--protein interaction networks via convex optimization,''
  \emph{Journal of Computational Biology}, no.~11, 2016.

\bibitem{zanfir2018deep}
A.~Zanfir and C.~Sminchisescu, ``Deep learning of graph matching,'' in
  \emph{2018 {IEEE} Conference on Computer Vision and Pattern Recognition,
  {CVPR} 2018, Salt Lake City, UT, USA, June 18-22, 2018}.\hskip 1em plus 0.5em
  minus 0.4em\relax {IEEE} Computer Society, 2018, pp. 2684--2693.

\bibitem{xu2019gromov}
H.~Xu, D.~Luo, H.~Zha, and L.~C. Duke, ``Gromov-wasserstein learning for graph
  matching and node embedding,'' in \emph{International conference on machine
  learning}.\hskip 1em plus 0.5em minus 0.4em\relax PMLR, 2019, pp. 6932--6941.

\bibitem{heimann2018regal}
M.~Heimann, H.~Shen, T.~Safavi, and D.~Koutra, ``Regal: Representation
  learning-based graph alignment,'' in \emph{Proceedings of the 27th ACM
  international conference on information and knowledge management}, 2018, pp.
  117--126.

\bibitem{trung2020adaptive}
H.~T. Trung, T.~Van~Vinh, N.~T. Tam, H.~Yin, M.~Weidlich, and N.~Q.~V. Hung,
  ``Adaptive network alignment with unsupervised and multi-order convolutional
  networks,'' in \emph{Proc. of ICDE}.\hskip 1em plus 0.5em minus 0.4em\relax
  IEEE, 2020.

\bibitem{gao2021unsupervised}
J.~Gao, X.~Huang, and J.~Li, ``Unsupervised graph alignment with wasserstein
  distance discriminator,'' in \emph{Proceedings of the 27th ACM SIGKDD
  Conference on Knowledge Discovery \& Data Mining}, 2021.

\bibitem{tang2023robust}
J.~Tang, W.~Zhang, J.~Li, K.~Zhao, F.~Tsung, and J.~Li, ``Robust attributed
  graph alignment via joint structure learning and optimal transport,'' in
  \emph{2023 IEEE 39th International Conference on Data Engineering
  (ICDE)}.\hskip 1em plus 0.5em minus 0.4em\relax IEEE, 2023, pp. 1638--1651.

\bibitem{titouan2019optimal}
V.~Titouan, N.~Courty, R.~Tavenard, and R.~Flamary, ``Optimal transport for
  structured data with application on graphs,'' in \emph{International
  Conference on Machine Learning}.\hskip 1em plus 0.5em minus 0.4em\relax PMLR,
  2019, pp. 6275--6284.

\bibitem{memoli2011gromov}
F.~M{\'e}moli, ``Gromov--wasserstein distances and the metric approach to
  object matching,'' \emph{Foundations of computational mathematics}, 2011.

\bibitem{cuturi2013sinkhorn}
M.~Cuturi, ``Sinkhorn distances: Lightspeed computation of optimal transport,''
  in \emph{Advances in Neural Information Processing Systems 26: 27th Annual
  Conference on Neural Information Processing Systems 2013. Proceedings of a
  meeting held December 5-8, 2013, Lake Tahoe, Nevada, United States}, 2013,
  pp. 2292--2300.

\bibitem{kantorovich1942translocation}
L.~V. Kantorovich, ``On the translocation of masses,'' in \emph{Dokl. Akad.
  Nauk. USSR (NS)}, vol.~37, 1942, pp. 199--201.

\bibitem{su2017order}
B.~Su and G.~Hua, ``Order-preserving wasserstein distance for sequence
  matching,'' in \emph{2017 {IEEE} Conference on Computer Vision and Pattern
  Recognition, {CVPR} 2017, Honolulu, HI, USA, July 21-26, 2017}.\hskip 1em
  plus 0.5em minus 0.4em\relax {IEEE} Computer Society, 2017, pp. 2906--2914.

\bibitem{solomon2016entropic}
J.~Solomon, G.~Peyr{\'e}, V.~G. Kim, and S.~Sra, ``Entropic metric alignment
  for correspondence problems,'' \emph{ACM Transactions on Graphics (ToG)},
  no.~4, 2016.

\bibitem{genevay2018learning}
A.~Genevay, G.~Peyr{\'{e}}, and M.~Cuturi, ``Learning generative models with
  sinkhorn divergences,'' in \emph{International Conference on Artificial
  Intelligence and Statistics, {AISTATS} 2018, 9-11 April 2018, Playa Blanca,
  Lanzarote, Canary Islands, Spain}, vol.~84.\hskip 1em plus 0.5em minus
  0.4em\relax {PMLR}, 2018, pp. 1608--1617.

\bibitem{chen2020graph}
L.~Chen, Z.~Gan, Y.~Cheng, L.~Li, L.~Carin, and J.~Liu, ``Graph optimal
  transport for cross-domain alignment,'' in \emph{Proc. of ICML}, vol.
  119.\hskip 1em plus 0.5em minus 0.4em\relax {PMLR}, 2020, pp. 1542--1553.

\bibitem{schmitzer2013hierarchical}
B.~Schmitzer and C.~Schn{\"o}rr, ``A hierarchical approach to optimal
  transport,'' in \emph{International conference on scale space and variational
  methods in computer vision}.\hskip 1em plus 0.5em minus 0.4em\relax Springer,
  2013.

\bibitem{alvarez2018structured}
D.~Alvarez{-}Melis, T.~S. Jaakkola, and S.~Jegelka, ``Structured optimal
  transport,'' in \emph{International Conference on Artificial Intelligence and
  Statistics, {AISTATS} 2018, 9-11 April 2018, Playa Blanca, Lanzarote, Canary
  Islands, Spain}, vol.~84.\hskip 1em plus 0.5em minus 0.4em\relax {PMLR},
  2018, pp. 1771--1780.

\bibitem{chen2018optimal}
Y.~Chen, T.~T. Georgiou, and A.~Tannenbaum, ``Optimal transport for gaussian
  mixture models,'' \emph{IEEE Access}, 2018.

\bibitem{lee2019hierarchical}
J.~Lee, M.~Dabagia, E.~L. Dyer, and C.~J. Rozell, ``Hierarchical optimal
  transport for multimodal distribution alignment,'' in \emph{Proceedings of
  the 33rd International Conference on Neural Information Processing Systems},
  2019, pp. 13\,475--13\,485.

\bibitem{luo2022differentiable}
D.~Luo, H.~Xu, and L.~Carin, ``Differentiable hierarchical optimal transport
  for robust multi-view learning,'' \emph{IEEE Transactions on Pattern Analysis
  and Machine Intelligence}, vol.~45, no.~6, pp. 7293--7307, 2022.

\bibitem{yang2023hotnas}
J.~Yang, Y.~Liu, and H.~Xu, ``Hotnas: Hierarchical optimal transport for neural
  architecture search,'' in \emph{Proceedings of the IEEE/CVF Conference on
  Computer Vision and Pattern Recognition}, 2023.

\bibitem{frogner2015learning}
C.~Frogner, C.~Zhang, H.~Mobahi, M.~Araya, and T.~A. Poggio, ``Learning with a
  wasserstein loss,'' \emph{Advances in neural information processing systems},
  vol.~28, 2015.

\bibitem{chizat2018scaling}
L.~Chizat, G.~Peyr{\'e}, B.~Schmitzer, and F.-X. Vialard, ``Scaling algorithms
  for unbalanced optimal transport problems,'' \emph{Mathematics of
  Computation}, vol.~87, no. 314, pp. 2563--2609, 2018.

\bibitem{fatras2021unbalanced}
K.~Fatras, T.~S{\'e}journ{\'e}, R.~Flamary, and N.~Courty, ``Unbalanced
  minibatch optimal transport; applications to domain adaptation,'' in
  \emph{International Conference on Machine Learning}.\hskip 1em plus 0.5em
  minus 0.4em\relax PMLR, 2021, pp. 3186--3197.

\bibitem{YangU19}
K.~D. Yang and C.~Uhler, ``Scalable unbalanced optimal transport using
  generative adversarial networks,'' in \emph{International Conference on
  Learning Representations}, 2019.

\bibitem{caetano2009learning}
T.~S. Caetano, J.~J. McAuley, L.~Cheng, Q.~V. Le, and A.~J. Smola, ``Learning
  graph matching,'' \emph{IEEE transactions on pattern analysis and machine
  intelligence}, vol.~31, no.~6, pp. 1048--1058, 2009.

\bibitem{chowdhury2019gromov}
S.~Chowdhury and F.~M{\'e}moli, ``The gromov--wasserstein distance between
  networks and stable network invariants,'' \emph{Information and Inference: A
  Journal of the IMA}, vol.~8, no.~4, pp. 757--787, 2019.

\bibitem{xu2019scalable}
H.~Xu, D.~Luo, and L.~Carin, ``Scalable gromov-wasserstein learning for graph
  partitioning and matching,'' in \emph{Proceedings of the 33rd International
  Conference on Neural Information Processing Systems}, 2019, pp. 3052--3062.

\bibitem{villani2009optimal}
C.~Villani \emph{et~al.}, \emph{Optimal transport: old and new}.\hskip 1em plus
  0.5em minus 0.4em\relax Springer, 2009, vol. 338.

\bibitem{borgwardt2005shortest}
K.~M. Borgwardt and H.-P. Kriegel, ``Shortest-path kernels on graphs,'' in
  \emph{Fifth IEEE international conference on data mining (ICDM'05)}.\hskip
  1em plus 0.5em minus 0.4em\relax IEEE, 2005, pp. 8--pp.

\bibitem{shervashidze2009efficient}
N.~Shervashidze, S.~Vishwanathan, T.~Petri, K.~Mehlhorn, and K.~Borgwardt,
  ``Efficient graphlet kernels for large graph comparison,'' in
  \emph{Artificial intelligence and statistics}.\hskip 1em plus 0.5em minus
  0.4em\relax PMLR, 2009, pp. 488--495.

\bibitem{shervashidze2011weisfeiler}
N.~Shervashidze, P.~Schweitzer, E.~J. Van~Leeuwen, K.~Mehlhorn, and K.~M.
  Borgwardt, ``Weisfeiler-lehman graph kernels.'' \emph{Journal of Machine
  Learning Research}, vol.~12, no.~9, 2011.

\bibitem{xu2020gromov}
H.~Xu, ``Gromov-wasserstein factorization models for graph clustering,'' in
  \emph{Proceedings of the AAAI conference on artificial intelligence},
  vol.~34, no.~04, 2020, pp. 6478--6485.

\bibitem{lacoste2016convergence}
S.~Lacoste-Julien, ``Convergence rate of frank-wolfe for non-convex
  objectives,'' \emph{arXiv preprint arXiv:1607.00345}, 2016.

\bibitem{scetbon2022linear}
M.~Scetbon, G.~Peyr{\'{e}}, and M.~Cuturi, ``Linear-time gromov wasserstein
  distances using low rank couplings and costs,'' in \emph{International
  Conference on Machine Learning, {ICML} 2022, 17-23 July 2022, Baltimore,
  Maryland, {USA}}, vol. 162.\hskip 1em plus 0.5em minus 0.4em\relax {PMLR},
  2022, pp. 19\,347--19\,365.

\bibitem{xie2021a}
Y.~Xie, Y.~Mao, S.~Zuo, H.~Xu, X.~Ye, T.~Zhao, and H.~Zha, ``A hypergradient
  approach to robust regression without correspondence,'' in
  \emph{International Conference on Learning Representations}, 2021.

\bibitem{zhang2016final}
S.~Zhang and H.~Tong, ``Final: Fast attributed network alignment,'' in
  \emph{Proceedings of the 22nd ACM SIGKDD international conference on
  knowledge discovery and data mining}, 2016, pp. 1345--1354.

\bibitem{zhong2012comsoc}
E.~Zhong, W.~Fan, J.~Wang, L.~Xiao, and Y.~Li, ``Comsoc: adaptive transfer of
  user behaviors over composite social network,'' in \emph{Proceedings of the
  18th ACM SIGKDD international conference on Knowledge discovery and data
  mining}, 2012, pp. 696--704.

\bibitem{yang2016revisiting}
Z.~Yang, W.~Cohen, and R.~Salakhudinov, ``Revisiting semi-supervised learning
  with graph embeddings,'' in \emph{International conference on machine
  learning}.\hskip 1em plus 0.5em minus 0.4em\relax PMLR, 2016, pp. 40--48.

\bibitem{zitnik2017predicting}
M.~Zitnik and J.~Leskovec, ``Predicting multicellular function through
  multi-layer tissue networks,'' \emph{Bioinformatics}, no.~14, 2017.

\bibitem{zhou2018deeplink}
F.~Zhou, L.~Liu, K.~Zhang, G.~Trajcevski, J.~Wu, and T.~Zhong, ``Deeplink: A
  deep learning approach for user identity linkage,'' in \emph{IEEE INFOCOM
  2018-IEEE conference on computer communications}.\hskip 1em plus 0.5em minus
  0.4em\relax IEEE, 2018.

\bibitem{du2019joint}
X.~Du, J.~Yan, and H.~Zha, ``Joint link prediction and network alignment via
  cross-graph embedding,'' in \emph{Proceedings of the Twenty-Eighth
  International Joint Conference on Artificial Intelligence, {IJCAI} 2019,
  Macao, China, August 10-16, 2019}.\hskip 1em plus 0.5em minus 0.4em\relax
  ijcai.org, 2019, pp. 2251--2257.

\end{thebibliography}

\begin{IEEEbiography}[{\includegraphics[width=1in,height=1.25in,clip,keepaspectratio]{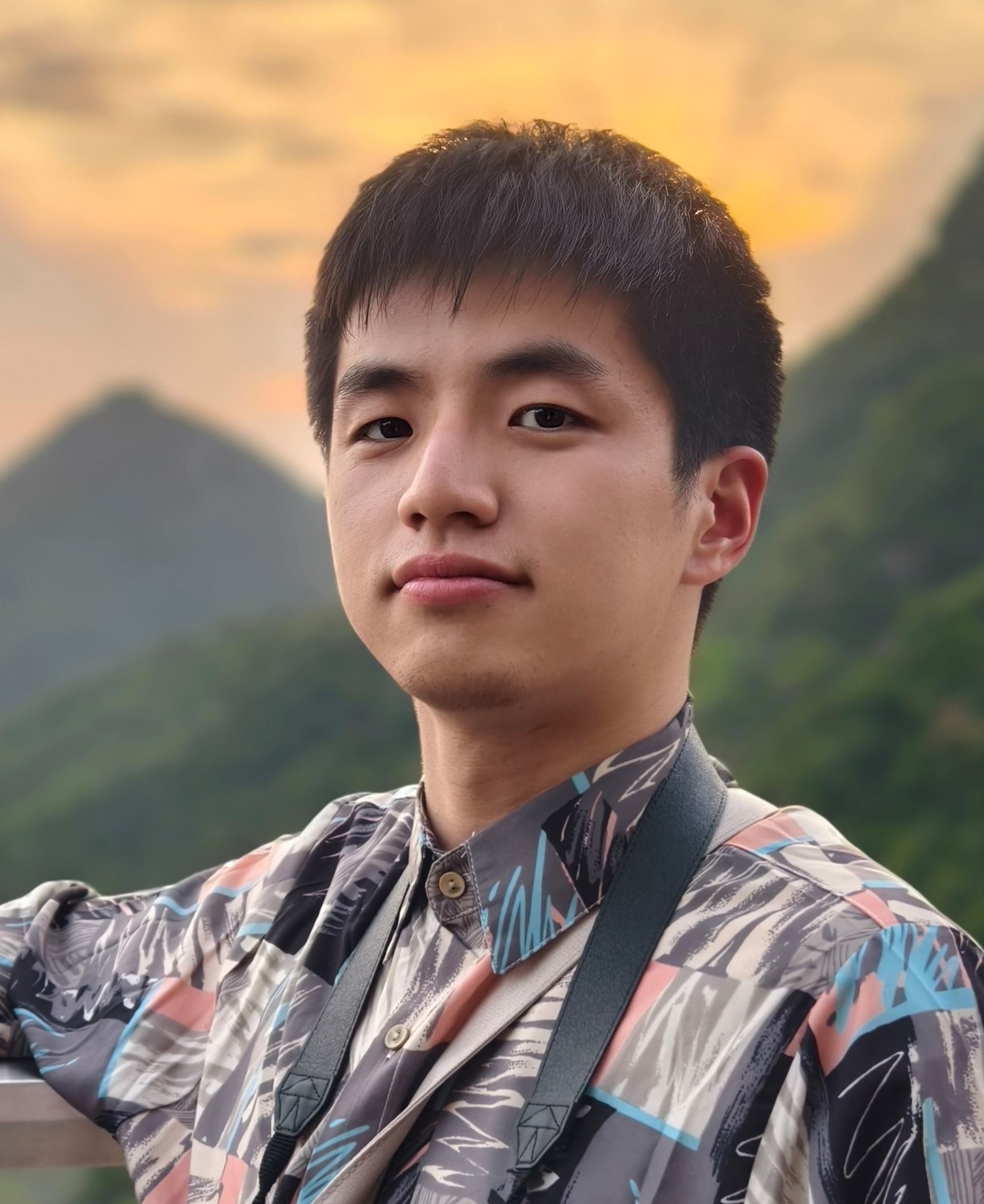}}]{Haoran Cheng}
received his B.S. degree in Computer Science from Beijing Institute of Technology, Beijing, China, in 2022, where he is currently pursuing an M.S. degree. 
His current research interests lie in machine learning and its applications, especially graph analysis.
\end{IEEEbiography}
\vspace{-3mm}
\begin{IEEEbiography}[{\includegraphics[width=1in,height=1.25in,clip,keepaspectratio]{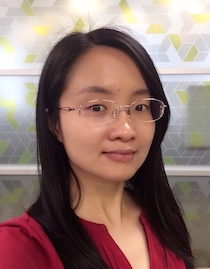}}]{Dixin Luo} is an Assistant Professor at the School of Computer Science and Technology at the Beijing Institute of Technology. 
From 2016 to 2020, she worked as a postdoctoral researcher at the University of Toronto and Duke University. 
She obtained her bachelor’s and PhD degrees from Shanghai Jiao Tong University in 2010 and 2016, respectively. 
Additionally, she was a visiting scholar at the School of Electrical and Computer Engineering at Georgia Institute of Technology from 2013 to 2014. 
Her research interests include machine learning and its applications to computer vision, sequential data analysis, graph analysis, and healthcare.
\end{IEEEbiography}
\vspace{-3mm}
\begin{IEEEbiography}[{\includegraphics[width=1in,height=1.25in,clip,keepaspectratio]{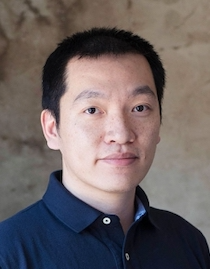}}]{Hongteng Xu} is an Associate Professor at the Gaoling School of Artificial Intelligence, Renmin University of China. 
From 2018 to 2020, he was a senior research scientist at Infinia ML Inc. 
During the same period, he was a visiting faculty member in the Department of Electrical and Computer Engineering, at Duke University. 
He received his Ph.D. from the School of Electrical and Computer Engineering at Georgia Institute of Technology (Georgia Tech) in 2017. 
His research interests include machine learning and its applications, especially optimal transport theory, sequential data modeling and analysis, deep learning techniques, and their applications in computer vision and data mining.
\end{IEEEbiography}
\end{document}